# Automated Linear-Time Detection and Quality Assessment of Superpixels in Uncalibrated True- or False-Color RGB Images

Andrea Baraldi[a,b,*], Dirk Tiede[b], and Stefan Lang[b]

[a] Department of Agricultural and Food Sciences, University of Naples Federico II, Portici (NA), Italy.
[b] Department of Geoinformatics – Z_GIS, University of Salzburg, Salzburg 5020, Austria.

* Corresponding author. Email: andrea6311@gmail.com

***Abstract*** — In this methodological paper, provided with a relevant survey value, an original low-level computer vision (CV) software pipeline, called RGB Image Automatic Mapper™ (RGBIAM™), is presented and discussed. RGBIAM is a lightweight computer program capable of automated near real-time superpixel detection and quality assessment in an uncalibrated monitor-typical red-green-blue (RGB) image, depicted in either true- or false-colors. The RGBIAM system design consists of known CV modules, constrained by the Calibration/Validation (*Cal/Val*) requirements of the Quality Assurance Framework for Earth Observation (QA4EO) guidelines. In agreement with the QA4EO *Cal* requirements, to benefit from multi-source data harmonization and interoperability, RGBIAM requires as mandatory an uncalibrated RGB image pre-processing first stage, consisting of an automated (self-organizing) statistical model-based color constancy algorithm. The RGBIAM's *hybrid* (combined deductive/top-down and inductive/bottom-up) inference pipeline comprises: (I) a direct quantitative-to-nominal (QN) RGB variable transform, where RGB pixel values are mapped (quantized) onto a prior dictionary of color names, to be community-agreed upon in advance, equivalent to a static (non-adaptive to data) polyhedralization of the RGB cube. Prior color naming is the deductive counterpart of popular inductive vector quantization (VQ) algorithms, whose typical VQ error function to minimize is a root mean square error (RMSE). In the output multi-level color map domain, superpixels are automatically detected in linear time as connected sets of pixels featuring the same color label. (II) An inverse nominal-to-quantitative (NQ) RGB variable transform, where a superpixelwise-constant RGB image approximation is generated in linear time, which allows to assess a VQ error image in compliance with the QA4EO *Val* requirements. The hybrid direct and inverse RGBIAM-QNQ transform is: (i) general-purpose, i.e., data- and application-independent. (ii) Automated, i.e., it requires no user-machine interaction. In the hybrid RGBIAM pipeline, a deductive inference first stage, analogous to genotype, provides automatically inherently ill-posed inductive learning-from-data algorithms, equivalent to phenotype, with initial conditions. (iii) Near real-time, i.e., its computational complexity increases linearly with the image size. (iv) Implemented in tile streaming mode, to cope with massive images. As a proof-of-concept, a realization of the RGBIAM pipeline was tested on three RGB images acquired by different imaging sensors and acquisition platforms. Collected outcome and process quantitative quality indicators, including degree of automation, computational efficiency, VQ rate and VQ error, are consistent with theoretical expectations and reveal that the RGBIAM lightweight computer program is suitable for low-level CV mobile software applications specifically designed to run on web browsers and mobile devices, such as tablet computers, smartphones and Unmanned Aerial Systems (UASs) mounting low-weight consumer-level color cameras.

## I. Introduction

OUTPACED by the rate of collection of images and videos of ever-increasing quality and quantity, such as those acquired by novel generations of spaceborne and airborne Earth observation (EO) imaging sensors, in addition to images acquired by consumer-level color cameras mounted on mobile devices, including unmanned aerial vehicles (UAVs) [104], the computer vision (CV) and remote sensing (RS) communities appear unable to transform image big datasets into operational, comprehensive and timely information products, in compliance with the Quality Assurance Framework for Earth Observation (QA4EO) guidelines [20]. The overarching goal of the present methodological paper is to contribute to invert this negative trend by fostering *information-as-data-interpretation* capabilities of a low-level CV system in operating mode, where quantitative (numeric and unequivocal) variables in the image domain are automatically transformed into nominal (categorical and inherently equivocal) variables [21], eligible for use in symbolic human reasoning, typically mimicked by fuzzy logic [83].

Formally, a finite image is a function that assigns colors (e.g., coded by real numbers) to a finite, rectangular array of locations in space (e.g., coded by ordered pairs of integers) [101]. There is a vast CV literature dealing with low-level (pre-attentional) image feature extraction, image analysis and synthesis (coding and decoding), image pair similarity assessment, image-object segregation (segmentation), high-level (attentive) image understanding (classification), etc. [1]-[17]. Unfortunately, links of CV to biological vision remain extremely weak [5], [7]-[12]. On the one hand, human vision can be considered a huge puzzle with a lot of missing pieces to date [9], [10]. On the other hand, computational models of simple, complex and end-stopped cells located in the primal visual cortex of mammals have been proposed in the last 10 years [4], [5], [9]-[12], [18]. In general, "no claim is made about the pertinence or adequacy of CV models to human visual perception… This enigmatic situation arises because research and development in CV is often considered quite separate from research into the functioning of human vision. A fact that is generally ignored is that *biological vision is currently the only measure of the incompleteness of the current stage of CV and illustrates that the problem is still open to solution*" [19]. For example, if we require that a CV system should be able to predict perceptual effects, such as the well-known Mach bands illusion where bright and dark bands are seen at ramp edges, then the number of published vision models becomes surprisingly small [12]. A diffuse inconsistency of existing CV systems with human visual perception may partly explain why the CV and RS communities are being outpaced by the rate of collection of images of ever-increasing quality and quantity. For example, still now the percentage of EO images downloaded by users from the European Space Agency (ESA)'s EO image databases is estimated at about 10% or less. In addition, no EO Level-2 prototype product has ever been generated systematically at the ground segment. A typical definition of an EO Level 2 product encompasses a multi-spectral (MS) image corrected for atmospheric, adjacency and topographic effects into surface reflectance values, provided with a scene classification map [105], [106].

Since the first key principle of *accessibility* to sensory data required by the QA4EO guidelines has greatly improved in recent years, a low exploitation rate of EO big data to be systematically transformed into Level 2 products may be due to an ongoing inadequacy of the second QA4EO key principle, specifically, *suitability* of data, information processing methods and information products, subject to a quantitative quality assurance ($Q^2A$) policy to be community-agreed upon in advance [20]. To better understand their potential impact upon the RS community, the two QA4EO key principles of accessibility to and suitability of sensory data and data-derived information products can be related to the two information theories investigated by philosophical hermeneutics, specifically, quantitative/ unequivocal/ objective *information-as-thing* and qualitative/ equivocal/ subjective *information-as-data-interpretation* [21]. These two concepts of information correspond intuitively to the two well-known classes of variables, either quantitative/numeric or qualitative/nominal/categorical. Unequivocal *information-as-thing*, related to the popular Shannon's mathematical theory of data communication/transmission, irrespective of the meaning of the transmitted message [22], is "easier to cope with" than equivocal *information-as-data-interpretation,* where the message receiver has a pro-active role in the message interpretation process [21]. This double meaning of the word information is a major cause of equivocality at the root of information technology (IT) [23], [24]. It explains why to date the QA4EO requirement of making quantitative RS data accessible has been easier to deal with than assuring suitability of qualitative information products generated by RS data interpretation initiatives.

To increase the suitability of EO image value-adding processes and products available for operational, timely and comprehensive exploitation of EO image "big data" by a wide public of image service industries, providers and end users [20], our work pursues a holistic approach to the study of vision as part of cognitive science [21], [25]-[30], see Fig. 1. Cognition is the transformation of ever-varying sensations into stable percepts/concepts [1], [2], [30]. To fill the information gap from low-level sensory data to high-level classes of objects in a *world model* [1], [2], cognitive processes, including vision, are inherently difficult, which means ill-posed in the Hadamard sense [31], [77], i.e., NP-hard [102], [103]. To become better posed for solution, they require *a priori* knowledge in addition to sensory data [2], [32], [33]. In practice, cognitive systems are *hybrid* inference systems where bottom-up inductive learning-from-data capabilities, typically investigated by machine learning [32], [33], must be combined with top-down knowledge-by-rule inference, which is the traditional focus of attention of artificial intelligence [6], [34]. A multidisciplinary cognitive approach to vision agrees with an increasing trend in scientific literature, where hybrid inference systems combine bottom-up statistical models with top-down physical models to take advantage of the unique features of each and overcome their shortcomings [23], [24], [35]. For the sake of completeness, pros and cons of deductive and inductive inference systems are summarized below.

- Relying exclusively on *a priori* knowledge available in addition to data, expert systems are static (non-adaptive to data) non-iterative (one-pass) decision trees that require neither user-defined parameters nor training data to run. Hence, they are called fully automatic. Their computation time is linear with the number of decision rules. In general, it takes a long time for human experts to learn physical laws of the real world-through-time and tune physical models. In addition, expert systems suffer from an intrinsic lack of flexibility, i.e., decision rules do not adapt to changes in the input dataset and in user needs, hence their knowledge base may soon become obsolete [34]. Finally, they suffer from an intrinsic lack of scalability, in particular static rule-based systems are impractical for complex problems. Nevertheless, once a syntactic inference system is set up and proved to be robust to changes in input data, then the effort pays off [23].
- Inherently ill-posed in the Hadamard sense [31], inductive learning-from-data algorithms require *a priori* knowledge in addition to data to become better posed for numerical solution [32]. It means that, in general, inductive data learning algorithms are semi-automatic and data-specific [35]. They are unable to learn class grammars and semantic networks of concepts as nodes and inter-concept relations, such as part-of, subset-of, etc., as arcs between nodes [6]. For training and testing phases, they require an adequate reference dataset whose quality, availability and costs can be impossible or unaffordable to cope with, especially when ground truth must be collected across time and geographical space, such as in EO data mapping problems at global scale.

Quite strikingly, instead of focusing on hybrid CV solutions, the mainstream CV and RS communities concentrate their research and technological development (RTD) efforts on driven-without-knowledge inductive learning-from-data algorithms, such as support vector machines [36], random decision forest classifiers [37] and the increasingly popular deep convolutional neural networks [38], [39], whose free-parameters to be user-defined based on heuristic criteria include the network architecture [32], [33].

In the multidisciplinary scenario of cognitive science, sketched in Fig. 1, the goal of the present low-level vision software design and implementation project was to develop an off-the-shelf (ready-for-use) hybrid reversible (direct and inverse) quantitative-to-nominal-to-quantitative (QNQ) transform of an uncalibrated monitor-typical red-green-blue (RGB) image for automated *superpixel* detection and $Q^2A$ in linear time. To become better conditioned for numerical solution, the low-level CV software pipeline was constrained by the following process and outcome requirements.

(I) To comply with the QA4EO Calibration/Validation (*Cal*/*Val*) requirements [20]. The QA4EO *Val* principle requires an EO data processing pipeline to provide each processing stage with a set of community-agreed quantitative quality indicators ($Q^2$Is), so that error propagation across the pipeline can be monitored in comparison with quality reference standards [20]. EO data *Cal* is the transformation of dimensionless digital numbers into a physical unit of radiometric measure, e.g., at-sensor reflectance, based on radiometric *Cal* metadata parameters [20]. In line with the QA4EO recommendations the RS community should regard as an indisputable fact that "the prerequisite for physical model-based, quantitative analysis of airborne and satellite sensor measurements in the optical domain is their calibration to spectral radiance" ([105]; p. 29). In general, availability of physical variables is a necessary not sufficient condition for sensory data analysis with physical models [6], [32], [35]. In addition to physical models, physical variables can be investigated by statistical models [6], [32], [35]. Hence, physical variables can be analyzed by hybrid inference systems. On the other hand, quantitative variables provided with no physical unit of measure can be investigated by statistical models exclusively [35]. Irrespective of this common knowledge, EO data *Cal* is largely oversighted in the RS common practice [23], [24].

(II) To run without user-machine interaction, i.e., it requires neither user-defined parameters nor training data to run. Inference system automation can be accomplished by a hybrid inference system where a deductive inference first stage employs *a priori* knowledge, available in addition to data, to provide second-stage inference learning-from-data algorithms with initial conditions without user-machine interaction. This hybrid inference approach complies with: (a) biological cognitive systems, where "there is never an absolute beginning" [29] because top-down genotype initializes bottom-up phenotype [30]. (b) The principle of statistical stratification. Well known in statistics, it states that "stratification will always achieve greater precision provided that the strata have been chosen so that members of the same stratum are as similar as possible in respect of the characteristic of interest" [40].

(III) To detect color-homogeneous texture elements, traditionally known as *texels*/ *textons* [6], [14]-[17], otherwise called *tokens* in agreement with the Marr definition of low-level vision's raw primal sketch [4]. Color is the sole quantitative visual information available at the pixel resolution, i.e., color values are context-insensitive, whereas all remaining visual properties, such as texture (perceptual spatial grouping of texels), image-object shape and size, inter-object spatial relationships, etc., are spatial-context dependent. In general, both in the spatiotemporal 4D real world-through-time domain, described in user-speak by a world model [1], [2], and in the (2D) image domain described in techno-speak, spatial information dominates color information, irrespective of data dimensionality reduction from 4D to 2D [102], [103]. In fact, human panchromatic vision is nearly as effective as human color vision in the provision of a complete scene-from-image representation in our brain: from local syntax of individual objects to global gist and layout of objects in space, including semantic interpretations and even emotions [5], [9], [10], [13]. Since spatial reasoning is required in high-resolution image understanding where spatial information dominates color information, the geographic object-based image analysis (GEOBIA) paradigm has become increasingly popular in the RS community [50], while the CV community has increasingly adopted semi-automatic inductive data learning algorithms for color-homogeneous *superpixel* detection as a prior step in many high-level CV tasks [41]. To the best of these authors' knowledge, the concept of superpixel is not novel, but highly related to the Julesz's

texton/ texel theory of pre-attentive human vision developed back in the 1970's [6], [14]-[17]. If this conjecture holds, terms superpixel and texel/ texton are synonyms and related to the Marr's raw primal sketch in pre-attentional vision [4].

(IV) To be general-purpose, i.e., data-, user- and application-independent. Selected by a user according to his/her own photointerpretation purposes, any true- or false-color three-band image can be uploaded onto a monitor-typical RGB cube. False-color image composites include 3-channel stacks of heterogeneous and/or dimensionless scalar 2-D gridded variables, such as image-derived spectral indexes (SIs, where each SI is equivalent to a 2-D membership function), such as a bare soil SI, a vegetation SI and a water SI uploaded onto the RGB channel R, G and B respectively [35].

(V) To run in linear time, i.e., the system complexity is O(*N*), where *N* is the image size.

(VI) To be implemented in tile streaming mode, to cope with massive digital images by reducing central memory occupation.

These constraints make the proposed low-level vision RGB image QNQ transform a lightweight computer program eligible for use in a mobile software application. By definition a mobile software application, eventually provided with a mobile user interface, is a lightweight computer program specifically designed to run on web browsers and mobile devices, such as tablet computers and smartphones.

If existing image segmentation algorithms are categorized according to the well-known hierarchical vision theory developed in the late 1970's by Marr [4], there is no rationale in comparing algorithms belonging to different information processing stages of the visual system pipeline. For example, the proposed automatic and data-independent low-level vision software system for superpixel/ texton/ texel detection fully complies with the Marr definition of raw primal sketch [4]. As a consequence, it is preliminary not alternative to any inductive semi-automatic and site-specific algorithm [35] for either: (a) texture segmentation (perceptual grouping) [1], [2], [13], equivalent to the Marr's low-level vision full primal sketch [4], where texture is the visual effect generated by the spatial distribution of texels detected at the raw primal sketch, or (b) surface segmentation, equivalent to the Marr's high-level 2½D sketch [4], e.g., refer to [42]. In addition to being inherently site-specific and semi-automatic, inductive data learning algorithms for texture segmentation or surface segmentation must be spatially context-sensitive, which boosts their computation time [13]. For example, in [42] spatial coherence is taken into account. Noteworthy, inherent functional properties of inductive data learning algorithms do not meet the aforementioned CV software project requirements.

Although the proposed low-level vision RGB image QNQ transform was developed within the RS community, its application domain extends to RGB images acquired by the whole CV discipline, which includes RS applications as a special case, see Fig. 1. Such an extended application domain of the proposed RGB image QNQ transform is a value added, i.e., it does not decrease the potential impact of the proposed solution upon EO image understanding applications.

The rest of this methodological paper is organized as follows. Section II provides this paper with a relevant survey value by presenting the problem background. Methods selected and proposed are discussed in Section III. Materials adopted in the experimental session are described in Section IV. Experimental results are presented in Section V and discussed in Section VI. Conclusions are reported in Section VII.

## II. Problem Background

### A. Human vision

The primary objective of any biological or artificial visual system is to back-project the information in the 2-dimensional (2D) image domain to that in the 3D viewed-scene domain, equivalent to a subset of the 4D real world-through-time observed at a given time [1], [2], [4]. To provide one or multiple plausible symbolic description(s) of a 3D viewed-scene, an image understanding system (IUS) is expected to find associations between ever-varying 2D quantitative (sub-symbolic) image features and stable symbolic concepts of 4D objects-through-time, e.g., houses, cats, etc. A finite and discrete ensemble of classes of 4D objects form a so-called 4D *world model* [2], or spatiotemporal ontology of the world-through-time [43]. It can be graphically represented as a semantic network, with nodes as classes of objects and arcs between nodes as inter-class relationships [2], [44]. This definition of vision means that image understanding is an ill-posed problem in the Hadamard sense [31]. First, vision is affected by data dimensionality reduction from 4D to 2D causing, for example, image occlusion phenomena, which are seamlessly filled in by human visual perception. Second, there is a well-known semantic information gap from sub-symbolic image features, either 0D points, 1D lines or 2D polygons [45], and plausible symbolic description(s) of a 3D viewed-scene. This is the same information gap thoroughly investigated by both philosophy and psychophysical studies of perception in cognitive science [2], [30]. In vision, we are always seeing objects we have never seen before at the sensation level, while we perceive familiar objects everywhere at the perception level [2]. In practice, the human visual system is able to construct on the basis of a brief glance a complete scene-from-image representation in our brain: from local syntax of individual objects to global gist and layout of objects in space, including semantic interpretations and even emotions [5].

To accomplish its cognitive tasks, the visual system of mammals comprises a pre-attentional vision first phase and an attentional vision second phase, summarized as follows.

(1) Pre-attentive/low-level vision extracts picture primitives based on general-purpose image processing criteria independent of the scene under analysis. It acts in parallel on the entire image as a rapid (< 50 ms) scanning system to detect variations in simple visual properties [7]-[12]. In each hypercolumn of the primary visual cortex (PVC), there are end-stopped cells, in addition to simple- and complex-cells [5], [10]. While simple- and complex-cells are thought to accomplish line and edge extraction [9], end-stopped cells respond to image singularities, such as line/edge crossings, vertices of image-objects and end-points of line segments [5], [10]. According to Marr [4], pre-attentive vision consists of two phases. (a) A *raw primal sketch* for token extraction, where tokens are 0D image singularities, 1D contours and 2D segments. (b) A *full primal sketch*, also known as perceptual grouping [14]-[17], where texture detection is accomplished based on the spatial distribution of tokens/texels. Noteworthy, in the words of Marr, "vision goes symbolic almost immediately, right at the level of zero-crossing (first-stage primal sketch), without loss of information" ([4]; p. 343).

(2) Attentive/high-level vision operates as a careful scanning system employing a focus of attention mechanism based on end-stopped cells [10], [18]. Scene subsets, corresponding to a narrow aperture of attention, are observed in sequence and each step is examined quickly (20–80 ms) [7], [8]. The Marr high-level vision subsystem comprises an understanding of local surface properties, called *2½D sketch*, followed by the generation of one or several plausible symbolic descriptions of the 3D viewed-scene [4].

As reported in Section 1, human panchromatic vision is nearly as effective as human color vision in the provision of a complete scene-from-image representation in our brain: from local syntax of individual objects to global gist and layout of objects in space, including semantic interpretations and even emotions [5], [9], [10], [13]. On an *a posteriori* basis, this observation has two important implications. First, in the 4D real world-through-time, color information of 4D objects, e.g., cars and trees, is dominated by their spatiotemporal attributes, as properly stated by Adams *et al.* [93]. Second, the same consideration holds for a planar representation of the 4D world-through-time, where 2D spatial (contextual) information dominates color information. To cope with the dominant 2D spatial information in an image, the human visual system employs modular arrays of multiscale 2D local filters [13], capable of a topology-preserving mapping of an image onto a neural network [46]-[49].

In compliance with the Marr's vision theory [4], a hierarchical ontology of a 3D viewed-scene could be obtained by formalizing a top-down language which should address all possible 3D model representations, down to an understanding of local surface properties, down to a description of the image properties [4]. On the other hand, intuitive bottom-up image understanding starts from scratch with a bunch of pixels, to be partitioned into image-objects to be grouped according to inter-object spatial relations (layout), etc., in agreement with the increasingly popular object-based image analysis (OBIA) paradigm [50], whose final output product consists of symbolic image-objects in a geographic information system (GIS)-ready file format. Both top-down and bottom-up inference approaches are possible, but the human visual system employs them jointly all the time [5]. For example, Vecera and Farah proved that "image segmentation can be influenced by the familiarity of the shape being segmented… Results are consistent with the hypothesis that image segmentation is an interactive (hybrid) process, in which top-down knowledge partly guides lower level processing… If an unambiguous, yet unfamiliar, shape is presented, top-down influences are unable to overcome powerful bottom-up cues… While bottom-up cues are sometimes sufficient for processing, these cues do not act alone; top-down cues, on the basis of familiarity, also appear to influence perceptual organization" ([51]; p. 1294).

### *B. RGB Cube Partitioning into "Universal" Basic Colors Known in Advance*

In a color model, each color is addressed by a unique coordinate in a color coordinate system. Based on various guiding principles, there are several coordinate arrangements that generate, in turn, color models. Terms like color space, color coordinate system and color model are generally used as synonyms [52]. Typically, color coordinate systems are categorized as objective or subjective. The subjective category is also termed "perceptual" because in perceptual color spaces, such as the Munsell system, color distances are closer to visual differences perceived by a human observer. Among objective color systems, the hardware-oriented RGB color model is the most commonly employed in color monitors [52]. It consists of a Cartesian coordinate system where colors are a linear additive mixing of the three RGB primary colors. Its gamut approximates well the gamut of surface colors, as the gamut of surface colors is not far from being parallelepiped in form, which is inevitable given the non-convexity of the spectral locus [53]. This may explain why seven of the RGB cube's vertices coincide well with the foci of seven of the eleven Berlin and Kay's human basic color (BC) categories [25], specifically, black, white, red, green, blue, yellow and purple [53], see Fig. 2. Central to this consideration is Berlin and Kay's landmark study of color words in 20 human languages. On the basis of that study they claimed that the 'basic color terms of any given language are always drawn' from a universal inventory of eleven color names: black, white, gray, red, orange, yellow, green, blue, purple, pink and brown [25]. These perceptual BC categories are expected to be "universal", i.e., users are not required to learn a new color representation for ever-varying sensory datasets, but they can apply the same universal color representation independently of the image at hand [54]. To summarize, in addition to be hardware-oriented, the RGB cube is a "natural" coordinate system, quite consistent with human BCs. If the RGB cube axes are discretized into known quantization levels, i.e., when the forward discrete color map is known, than an efficient inverse color map algorithm can replace each input RGB pixel with the center value of its assigned discrete color cell [55]. In [52], the reversibility of six geometric color spaces was investigated, where reversibility was defined as the capacity of a color model to recover an original 8-bit depth RGB image representation in $(256)^3$ RGB combinations, after transforming the $(256)^3$ RGB combinations into the target color space and then transforming back to the RGB cube. The conclusion was that the reversibility error in the forward and backward color space transformations was negligible for human visual interpretation purposes.

*C. Color Constancy*

In human vision, color constancy ensures that the perceived color of objects remains relatively constant under varying illumination conditions, so that they appear identical to a "canonical" (reference) image, subject to a "canonical" (known) light source (of controlled quality), e.g., under a white light source [56]. In short, solution of the color constancy problem is the recovery "of an illuminant-independent representation of the reflectance values in a scene" [57]. In practice color constancy supports image harmonization and interoperability when no radiometric calibration parameter is available. Its goal is analogous to that of inter-image relative calibration [35] and image-specific absolute radiometric *Cal*, considered mandatory by the QA4EO guidelines when radiometric calibration parameters are available [20].

Color constancy is intrinsically related to brightness in a more global (large scale) sense than, say, image-contour detection, which depends on local non-stationary image properties. Although the biological mechanisms involved with the color constancy ability are not yet fully understood [3], [56], it is speculated that a special type of retinal ganglion cells can play a role in the estimation of a global "background" brightness, on which lines, ramps and step edges [58] can be projected [5]. In principle, this special type of retinal ganglion cells can be involved with brightness perception because: (i) it features a very large receptive field. (ii) It is not connected to either rods or cones. (iii) It is connected to central brain areas for controlling the circadian clock (day-night rhythm). (iv) Via a feedback loop, it is connected to the eye's iris (pupil size). (vi) These special retinal cells also connect to at least the ventral area of the lateral geniculate nucleus [5].

Computational color constancy is a fundamental prerequisite of many CV applications, such as the RGB image QNQ transform proposed in this paper. Also because biophysical mechanisms of color constancy remain largely unknown, computational color constancy algorithms are typically unable to simulate color constancy effects observed in humans [19]. Computational color constancy is an under-constrained problem in the Hadamard sense [31]. Since it does not have a unique solution, it requires *a priori* knowledge in addition to data for numerical treatment [32]. For these reasons there has been a large number of alternative color constancy algorithms proposed in the CV literature in the last 30 years [59]. Early computational models were derived from works on human perceptual theory, resulted in the Retinex perceptual theory by Land [60], considered inadequate by now. In survey works such as [3], [56] and [59], computational color constancy approaches are divided into three categories. (1) Low-level statistical model-based methods. (2) Physical model-based methods. (3) Gamut-based methods. In statistical models, the best-known statistical assumption about color distributions is the so-called *Grey-World* assumption: the average reflectance in a scene under a neutral light source is achromatic, which means that the color of the light source can be estimated by computing the average color in the image. Another well-known assumption is the *White-Patch* assumption: the maximum response in the RGB channels is assumed to be caused by a perfect reflector. The assumption of perfect reflectance is alleviated by considering the color channels separately, resulting in the *max-RGB* assumption, where the illuminant is estimated as the maximum response in each color channel separately. In common practice, if an image contains few edges (corresponding to a "low" signal-to-noise ratio), then pixel-based (context-insensitive, 1st-order spatial distribution) methods, like Grey-World and White-Patch, are preferred. When the signal-to-noise ratio is "medium" or "high", context-sensitive (edge-based) methods are preferred: for example, 1st- and 2nd-order local spatial derivatives are adopted in the so-called Grey-Edge method [61]. In [59], the eleven "universal" human BC categories, identified by Berlin and Kay [25], see Fig. 2, were adopted as a form of *a priori* knowledge, available in addition to and independent of data (refer to Section II.B), to improve the computational efficiency of the color-by-correlation algorithm proposed in [57]. In the statistical color constancy algorithm implemented in the Environment for Visualizing Images (ENVI) commercial software toolbox [62], "ENVI does a special ('ultimate') stretch for the display case, which really can't be reproduced using the ENVI's default 2% stretch. If the histogram features more than three bins, the special stretch will calculate the left hand percent stretch on hist[1:*] and the right hand percent stretch on hist[0:n_elements(hist)-2]. If the histogram is a Gaussian (normal) shaped curve, then the difference between this and the 'full' histogram is negligible. However, if there is a large saturation of min or max values (such as an image with a lot of background, e.g., water pixels, and/or foreground, e.g., cloud pixels), then ENVI's default stretch will ignore the spike(s) and calculate the percent linear stretch, e.g., a 2% stretch, based on the rest of the "real" histogram. This allows ENVI to display, by default, many images which otherwise would not stretch well with a 2% linear stretch since they contain more than 2% background and/or foreground" [63].

For a complete survey of computational color constancy methods, the interested reader can refer to the existing literature, e.g., [3], [56], [59].

### D. Deductive and Inductive Vector Quantization

In his seminal work conceived to bridge independent studies on color naming conducted by linguistics and CV [53], Griffin verified the hypothesis that the best system of color categories for pragmatic purposes coincides with human BCs, see Fig. 2. In line with [53], additional relationships among independent studies conducted by linguistics, inductive machine learning and deductive artificial intelligence can be identified to shed new light on the problem of prior knowledge-based color space partitioning/ discretization/ quantization. In the multidisciplinary framework of cognitive science (see Fig. 1), first, prior knowledge-based color space discretization is equivalent to color naming in a natural language [3], [64], [65]. According to linguistics, a discrete and finite set of BC names (refer to Section II.B) must be community-agreed upon in advance to become "universal", which means consistently used by members of the community on a regular basis [53

Second, prior knowledge-based color space quantization is the deductive counterpart of inductive unlabeled data learning algorithms for vector quantization (VQ) [32], [33] and data compression [52]. Two of the most popular and widely used VQ heuristics in unlabeled data analysis are the *k*-means VQ algorithm, also known as Lloyd's or Linde-Buzo-Gray's VQ algorithm [47], [66]-[70], and the Iterative Self-Organizing Data Analysis Technique (ISODATA) [71]. Typically, inductive VQ algorithms have to minimize a known VQ error function given a number of *k* discretization levels defined beforehand, such as the *k*-means VQ [47], whose practical time complexity is equal to $O(N \cdot k \cdot I)$ [72], [73], where variable $I$ is the number of iterations required to reach convergence. Vice versa, a user can fix the target VQ error value, so that it is the parameter *k* to be dynamically learned from unlabeled data by the inductive VQ algorithm [66], [67], such as ISODATA [71].

It is worth mentioning here that unsupervised data learning VQ algorithms should never be confused with unsupervised data clustering algorithms [32]. The goal of unsupervised data clustering is to locate hidden "perceptual" (fuzzy) data structures (clusters, hypervolumes) of any possible shape in the unlabeled data set at hand [46]-[49]. Unsupervised data clustering is a non-predictive unlabeled data mapping task [32], i.e., it is expected to perform the best with the available input dataset, irrespective of possible unknown future samples. Unlike VQ error minimization problems, there is no known cost function to minimize in unsupervised data clustering [47]-[49]. Finally, to provide a topology-preserving map of each unlabeled data cluster, unsupervised data clustering algorithms model both centroids (centers of mass) of processing elements, similar to those estimated by inductive VQ algorithms, but also lateral connections (synapses) between pairs of processing elements [46], to form one connected network of processing elements per data cluster [47]-[49].

Back to deductive and inductive VQ algorithms for color space partitioning, when they are compared, their functional differences make them complementary rather than alternative in nature, as discussed below.

(1) *Degree of automation*, directly related to the ease of use and inversely related to the number of system's free-parameters to be user-defined. On the one hand, prior knowledge-based color space partitioning is automatic and data-independent, refer to Section I. On the other hand, inductive VQ algorithms are inherently ill-posed [32], semi-automatic and site-specific [35], refer to Section I. For example, the *k*-means VQ algorithm requires the following free-parameters to be user-defined based on heuristics: (i) the number of vector quantization levels *k*, (ii) a convergence threshold, typically defined as the maximum number *I* of training iterations, and (iii) a dictionary (codebook) of *k* vector data centroids (centers of mass) to be initialized, e.g., by means of random data sampling.

(2) *Measurement space partitioning*. In inductive VQ algorithms a popular VQ error function to minimize is the root mean square error [32], [33], $RMSE_b$, $b = 1, \ldots, B$, defined as:

$$RMSE_b = \sqrt{\frac{\sum_{i=1}^{N}\left[P_b(i) - P_b^*(i)\right]^2}{N}}, \quad b = 1,\ldots, B, \qquad (1)$$

where *N* is the total number of pixels, $P_b(i)$ is the scalar value of the ith-pixel in band $b = 1, \ldots, B$, $P^*_b(i)$ is the post-quantization ith-pixel value, while the adopted metric distance is the Euclidean distance. When they adopt a Euclidean metric distance minimization criterion, such as Eq. (1), and they reach convergence, then inductive VQ algorithms, such as *k*-means and ISODATA, accomplish a Voronoi tessellation of the input data space, which is a special case of convex polyhedralization [46], [47]. On the contrary, the designer of a prior knowledge-based decision tree for color space discretization is free to adopt discretization levels of any possible shape and size, either convex or concave, either connected or not, see Fig. 2. Unfortunately, when the MS space dimensionality is superior to three, a prior dictionary of mutually exclusive and totally exhaustive hyper-polyhedra is difficult to think of and impossible to visualize.

(3) *Computational complexity*. Because color is the sole visual information to be context-independent [1], [2], [4] (refer to Section I), then a color space discretization algorithm, either inductive or deductive, is expected to work with pixel values as vector data. A pixel-based VQ considers a (2D) image as a 1D vector stream, featuring no spatial information. In this case, an iterative suboptimal *k*-means VQ algorithm has a practical time complexity equal to $O(N \cdot B \cdot k \cdot I)$ [72], [73], where variable *I* is the number of iterations required for convergence and *B* is the number of color channels. On the contrary, an expert system for color space discretization is a one-pass static decision tree, with one decision rule per target BC category. Hence, its complexity is equal to $O(N \cdot B \cdot BC)$, where *BC* is the number of basic color categories known *a priori*.

Based on the aforementioned comparison, in addition to considerations reported in Section I about the complementary nature of inductive and deductive inference, it is important to conclude that *deductive and inductive VQ algorithms should never be considered alternative, but complementary in nature*. For example, a deductive color space quantization first stage can be employed to initialize automatically, based on prior color knowledge, the number *k* of vector data centroids and their initial values in an inductive driven-without-knowledge *k*-means VQ algorithm. The resulting hybrid inference system would be an automatic driven-by-knowledge *k*-means VQ algorithm. A realization of this hybrid inference concept can be found in [74], where an expert system for RGB cube partitioning was employed to initialize the semi-automatic and site-specific inductive data learning algorithm for image segmentation [75] adopted by the popular eCognition commercial software product [76] for OBIA applications [50].

### *E. Two-Pass Connected-Component Multi-Level Image Labeling*

Let's define: (i) an image as a 2D gridded quantitative/numeric variable [17] (refer to Section 1), and (ii) a multi-level image as a 2D gridded qualitative/categorical/nominal variable. Image segmentation [42], [75] is the dual problem of image contour detection [77]. These are both inherently ill-posed problems in the Hadamard sense [23], [24], [31], [75], [77]. They admit no unique solution and require *a priori* knowledge in addition to data to become better posed for numerical treatment [32]. On the contrary, a multi-level image, such as a classification map, can be deterministically partitioned into connected image-objects, consisting of 0D, 1D or 2D planar objects [45]. This is a well-posed planar segmentation problem, whose deterministic solution is unique. Whereas exactly one segmentation map can be derived from one multi-level image, the vice versa does not hold, i.e., the same segmentation map can be extracted from different multi-level images [78], see Fig. 3. Unfortunately, well-posed multi-level image segmentation algorithms are often confused with inductive ill-posed image segmentation algorithms. For example, a two-pass connected-component multi-level image labeling algorithm is automatic and its two-pass computational complexity $O(2 \cdot N)$ is linear in the image size in pixels, $N$ [6], [79]. In addition, to cope with massive images by reducing central memory occupation, it can be implemented in tile streaming mode [6], [79].

### *F. Superpixel Detection Equivalent to Texel Detection in the Pre-Attentional Raw Primal Sketch*

To the best of these authors' knowledge the concept of superpixel, developed by the CV community in recent years, is equivalent to the Julesz's texton/texel theory of pre-attentive human vision [6], [14]-[17] (refer to Section I). In CV, superpixel detection is an image pre-processing first stage employed for image simplification as input to high-level vision tasks. It is required to be fast to compute, memory efficient, simple to use by featuring few and intuitive input parameters to be user-defined, and capable of increasing the speed and quality of the higher-level vision tasks [41]. One popular inductive algorithm for superpixel detection is the simple linear iterative clustering (SLIC) [41], which is an adaptation of the popular $k$-means VQ algorithm [47], [66]-[70], refer to Section II.D. The SLIC algorithm's free-parameters to be user-defined based on heuristics are $k$, the desired number of approximately equally-sized superpixels, and $m$, a compactness term in range [1, 40], set by default equal to 10. If parameter $m$ increases, then detected superpixels tend to feature more regular size and shape [41]. The complexity of SLIC is linear in the total number of pixels $N$, $O(N)$, irrespective of $k$. According to the SLIC authors, "the (user's) ability to specify the amount of superpixels, and to control the compactness of the superpixels" are important [41]. This would agree with a popular criterion of *good* segmentation requiring that "a good segmentation region should be formed by connected pixels with homogeneous colors whose shape should be as compact as possible" [42], [80]. However, in common practice, the SLIC dependence on two unknown parameters to be user-defined based on empirical criteria decreases the algorithm's degree of automation (ease of use) and has a negative impact on its robustness to changes in input parameters and to changes in input data. For example, how can a user predict a reasonable number $k$ of equally-sized superpixels to be detected in massive images of complex real-world scenes, such as those tested in the further Section V? According to the present authors, by scoring low in degree of automation the inductive SLIC algorithm is little useful in common practice.

## III. METHODS

An original low-level vision software pipeline was implemented to accomplish a QNQ transform of a monitor-typical RGB image for superpixel detection and $Q^2A$, subject to the CV software project requirements listed in Section I. Specifically, the low-level CV software realization was required to be automatic, linear-time, data-, user- and application-independent, and in tile streaming mode. Our system realization was a proof-of-concept. It proved that the target project admits solution(s), based on existing algorithms, but it does not claim to be the "best" solution, if any exists. Actually, *information-as-data-interpretation* problems [21], such as cognitive problems including vision, are inherently equivocal/ subjective/ ill-posed (refer to Section I), i.e., there is no absolute "best" solution to *information-as-data-interpretation* problems [21].

### A. Software Design, Algorithm Selection and Implementation

The implemented RGB image analysis and synthesis software pipeline, consisting of six subsystems identified as block 1 to 6 in Fig. 4, is described below.

*RGB image analysis for superpixel/ texel detection, refer to blocks 1 to 5 in Fig. 4.*

1. *RGB image pre-processing for color constancy*. According to Section II.C, color constancy is considered mandatory to guarantee uncalibrated image harmonization and interoperability when no calibration metadata parameters are available, such as in UAVs employing low-weight consumer-level color cameras [104]. By analogy with the QA4EO *Cal* requirements [20], uncalibrated image color constancy is considered a necessary not sufficient condition for sensory data analysis with hybrid inference systems, where physical and statistical models are combined to take advantage of each and overcome their shortcomings. There is a wide variety of published algorithms for image color constancy [56], [57], [59], [61]. To comply with the software project requirements proposed in Section I, we designed and implemented an original self-organizing statistical algorithm (never published, patent pending) for $1^{st}$-order histogram-based (non-contextual) image color constancy in linear time $\leq O(I \cdot N \cdot B)$, where the number of learning-from-data iterations $I$ is $\leq 3$. It was inspired by the ENVI "ultimate" image stretching algorithm summarized in Section II.C. Our solution analyzes each single channel to detect one-of-four $1^{st}$-order histogram distributions, described as follows (see Fig. 5). (i) Neither a background nor a foreground mode is present in addition to a central mode. (ii) A background mode with a long right tail can be identified. (iii) A foreground mode with a long left tail can be identified. (iv) One background and one foreground mode can be identified in addition to a central mode. Once background and foreground modes are detected, if any, they are mapped onto the minimum output gray value, hist[0], and the maximum output gray value, hist[255], respectively. Next, a standard linear stretching algorithm is applied per channel, to fill the histogram bins hist[1:254], according to a traditional max-RGB criterion (refer to Section II.C).

2. *RGB cube partitioning into static (non-adaptive to data) polyhedra corresponding to BC names known a priori.* To bridge independent studies on color naming conducted by linguistics and CV, Griffin verified the hypothesis that the best system of color categories for pragmatic purposes coincides with human BCs [53], refer to Section II.B. By using a classification task to test this hypothesis, he obtained results consistent with it. In [53], the test dataset consisted of color RGB jpeg-format images collected by means of a web-based search-by-noun engine. A BC category system was generated with a multi-step process. First, the 267 Munsell coordinate-specified chips [81] were assigned with (gathered into) the eleven BC names. Next, color chips were mapped into the Commission Internationale de l'Eclairage (CIE)-Lab color space [52]. Finally, the eleven BC extents in CIE-Lab space were transformed into data expressed over a uniform 323 sampling of the monitor-typical RGB cube [52]. The final result was a hardware-oriented RGB cube partitioned into eleven mutually exclusive and totally exhaustive human-derived BC categories, as shown in Fig. 2.

In [54], if compared against inductive learning-from-data descriptors, static "universal" color descriptors are expected to cause a drop of quantization accuracy, counterbalanced by an increase in computational efficiency and degree of automation. Acknowledged that no single universal ("best") color dictionary exists, because any color naming is a conventional *information-as-data-interpretation* process to be community-agreed upon in advance [21], three alternative RGB cube partitions, featuring 11, 25 and 50 color clusters respectively, were investigated in [54]. Each color cluster gathered color cells that must be connected in the L*a*b* cube, starting from a total number of equally spaced color cells equal to m = 4000 = 10 × 20 × 20. In [54], no parent-child inter-cluster relationships exist at the different static color quantization levels.

We designed and implemented an original software solution for prior knowledge-based RGB cube partitioning in compliance with the project requirements listed in Section I. Called RGB Image Automatic Mapper™ (RGBIAM™, never published, patent pending), it found inspiration in the existing Satellite Image Automatic Mapper™ (SIAM™), an expert system for automatic transformation of a radiometrically calibrated EO multi-spectral image onto a set of color maps whose legends are color dictionaries featuring parent-child relationships [23], [24], [82]. In particular, RGBIAM is a one-pass prior knowledge-based decision tree for RGB cube partitioning into static polyhedra, non-adaptive to data and not necessarily convex and/or connected. Any prior knowledge-based decision tree encompasses a structural and a procedural knowledge. The former relates to the adopted set of decision rules, the latter to their order of presentation. By changing either its structural or procedural knowledge, the decision tree realization changes [23], [24]. Like in SIAM [82], the RGBIAM's decision rules define their individual domain of activation in the measurement space as one or more polyhedra, each one described by shape and intensity. First, it transforms each quantitative R, G and B variable into a qualitative variable consisting of fuzzy sets (FSs), e.g., low (L), medium-low (ML), medium-high (MH) and high (H), not necessarily uniform, according to the principles of fuzzy logic [83], [84]. Second, it identifies quantitative inter-channel relationships, called spectral rules (SRs), e.g., SR1 = max{B, G} < (0.5 * R). Third, it defines color names, called spectral categories (SC), as polyhedra combining SRs for shape and FSs for intensity, e.g., Color 1 = Bright Dominant Red = SR1 AND H_R. Two color discretization levels were implemented: (a) a fine color discretization level, consisting of 49+1 = 50 color names, including class "unknown", and (b) a coarse color discretization level, consisting of the 11 human BCs (refer to Section II.B) plus 1 class "unknown" = 12 color names, generated as a fixed parent-child combination of the 50 color names available at the fine discretization level, see Fig. 6. The RGBIAM's static decision tree computational complexity is ≤ $O(C1 \cdot N \cdot B + C2 \cdot N)$, where $C1 = 50$ = cardinality of the fine-granularity color dictionary and $C2 = 12$ = cardinality of the coarse-granularity color dictionary, generated as an aggregation of the $C1$ color names, i.e., inequality $C2 < C1$ must hold.

The (SIAM and) RGBIAM's output color maps are called image QuickMap™ products, as opposed to the traditional image QuickLook (typically, a true-color image in the jpg-file format) promoted by ESA in its user-driven EO image retrieval systems. Let's assume that the input RGB image is byte-coded, hence each pixel value requires 24 bits of memory space. An RGBIAM's map whose codebook consists of 50/12 color names requires 6/4 bits per pixel respectively. It means that RGBIAM works as an RGB data compression system at a given compression rate of 4:1 up to 6:1. In common practice, by featuring superior levels of data compression and semantics an image QuickMap can replace any QuickLook image employed in user-driven EO image retrieval systems.

3. *Well-posed extraction of connected components from a multi-level color map.* To transform a multi-level color map into a segmentation map, a well-posed two-pass connected component multi-level image labeling algorithm was implemented according to the existing literature [6], [79], refer to Section II.E. In our software pipeline, a two-scale segmentation map, where inter-scale image-objects feature parent-child relationships, was generated with computational complexity O(2·$N$·2), where a factor of 2 is due to the generation of two single-scale segmentation maps, one for each of the two RGBIAM's color maps featuring 50 and 12 color levels respectively. An original non-trivial tile streaming implementation of this algorithm was pursued to comply with the project requirements, refer to Section I.

4. *Well-posed contour extraction from known image-objects*. The dual problem of image segmentation is contour detection [23], [24], i.e., contours are image-object perimeters. When segments are detected beforehand, their deterministic (non-equivocal) contours can be coded in either raster or vector format. For example, non-equivocal contours of detected superpixels are shown in [41]. In compliance with the software project requirements listed in Section I, we extracted contours of detected superpixels in linear time and raster format by means of a well-posed one-pass 4- and 8-adjacency cross-aura estimate, implemented in tile streaming mode [85], see Fig. 7. The algorithm complexity is O((4·$N$ + 8·$N$)·2), where the factor 2 is due to the presence of two RGBIAM's color maps to extract contours from. Noteworthy, the 4-adjacency cross-aura measure is useful for high-level attentional OBIA applications [50], to be pursued in series with the RGB image QNQ transform, refer to processing block 7 in Fig. 4. For example, superpixel/texel contours can be used to partition an image into high- or low-texture areas, according to the empirical rule proposed in [1]. If a moving window of size $W \times W$ centered on pixel $p$ contains more than 2$W$ boundary pixels, then the central pixel $p$ can be marked as belonging to the high-texture image layer. Moreover, the 4-adjacency cross-aura allows estimation of a scale-invariant shape index of compactness, also called roundness (*Rndnss*) [7]. In particular,

$$Rndnss = (4 \times \text{sqrt}(A) / PL) \in [0, 1], \quad (2)$$

where $A$ is the segment area and $PL$ is the cumulative 4-adjacency cross-aura measure of the region's total boundary, where the total boundary takes into account contributions from inner holes, if any, i.e., total boundary = external (outer) boundary + inner boundary (due to holes). It can be easily proved (by induction) that this *Rndnss* formulation is scale invariant. For example, a 0D planar object consisting of a single pixel features $PL$ = 4, then *Rndnss* = 4/4 = 1 (maximum). For a 4-pixel square object, $PL$ = 8, then *Rndnss* = 4*2/8 = 1, etc. On the contrary, most of the existing formulations of *Rndnss* are not scale invariant in raster imagery [1], [2], [6], [86], [87].

5. *Segment description table allocation and initialization*. Positional, colorimetric, geometric and spatial attributes of raster image-objects can be stored in tabular format in a so-called segment description table (SDT) [1], to be dynamically allocated and initialized in central memory. For example, in their seminal works at the root of the OBIA paradigm [50], where they try to mimic the convergence-of-evidence approach adopted by human reasoning, Nagao and Matsuyama employed the tabular information of an SDT to classify image-objects based on converging colorimetric, geometric, textural and spatial evidence [1], [2]. Starting from the input RGB image and the two segmentation maps generated from the two RGBIAM's color maps of the input RGB image, the computational complexity of an SDT initialization phase is O(($N$·$B$)·2), with $B$ = 3, where the factor of 2 is due to the presence of two SDTs, one per segmentation map. To assess the central memory occupation of an SDT, as an example, let us consider a big RGB image, say, rows = $RW$ = columns = $CL$ = 50000, bands $B$ = 3, byte-coded, whose number of segments, detected by the RGBIAM expert system, is assumed to be equal to ($RW \times CL$ / 5 pixels per segment as average) = 5 × 10^8. The expected SDT memory occupation per segment would be the following. Segment identifier (ID) = unsigned long int = 4 bytes, locational property (minimum enclosing rectangle, defined by the upper right and lower left corners) =

unsigned long int × 4 = 16 bytes, RGBIAM's color label = unsigned char = 1 byte, area size = unsigned long int = 4 bytes, colorimetric mean = float × 3 bands = 12 bytes. Hence, the SDT memory occupation per segment is around 37 bytes. In this example, the SDT central memory occupation, equal to the number of segments × memory occupation per segment, would be 5 × 10^8 × 37 bytes = 18.5 GB. It means that, in general, the tabular representation of the raster image-object attributes in an SDT can be considered very demanding in terms of central memory occupation.

*RGB image synthesis (reconstruction) from the segmentation map and the SDT, refer to block 6 in Fig. 4.*

6. *Piecewise-constant input image approximation*. A one-pass piecewise-constant input image approximation, called "object-mean view" in the eCognition commercial software product [76], is equivalent to an edge-preserving smoothed image [1], such as that required in frames of a video sequence [88]. It was accomplished in near real-time by replacing each pixel scanned in the segmentation map with the SDT's colorimetric mean value of the superpixel that pixel belongs to. In the reconstruction of any RGB image typically characterized by non-stationary local statistics, e.g., mean, standard deviation, spatial autocorrelation, etc., an "object mean view" approach is expected to be more accurate, because more sensitive to varying local statistics, than an inverse color mapping where each input RGB pixel is replaced by the center value of its assigned discrete color cell, such as that proposed in [55]. The complexity of the implemented superpixelwise-constant image approximation algorithm is $O((N \cdot B) \cdot 2)$, with $B = 3$, where the factor 2 is due to the presence of two STDs, corresponding to the two detected color maps whose color codebook is $C1 = 50$ and $C2 = 12$ respectively. Since edge-preserving image smoothing follows image segmentation into superpixels, it is completely alternative to traditional edge-preserving image smoothing via 2D spatial filtering, which is required before image segmentation applied to frames of a video sequence [88]. For robotic applications it is very important that the labels of image segments do not change throughout a video stream. Currently only very few segmentation algorithms running in real-time achieve this objective, among them the Metropolis algorithm [89]. The conclusion is that the proposed automatic edge-preserving image smoothing filter is expected to be particularly useful in reducing oversegmentation phenomena that typically affect the Metropolis algorithm in textured areas of a video sequence [88].

*Summary of blocks 1 to 6 in Fig. 4.*

The 6-stage low-level vision software pipeline shown in Fig. 4 is implemented in tile streaming mode. Its overall computational complexity is $\leq O(N \cdot ((C1 \cdot B + 7 \cdot B) + C2 + 28))$, which is linear in the image size $N$, number of bands $B$ and number of color quantization levels $C1$ and $C2$, with $C2 < C1$. This software pipeline implementation complies with the software project requirements listed in Section I.

### B. Quantitative Quality Assurance ($Q^2A$) of the Low-level Vision Software Pipeline

For $Q^2A$ of the RGB image QNQ converter, a minimally redundant and maximally informative set of outcome and process quantitative quality indicators (OP-$Q^2$Is), encompassing both outcome $Q^2$Is and process $Q^2$Is, must be selected, to be community agreed-upon in advance, in compliance with the QA4EO guidelines [20]. To be considered operational, an information processing system must score "high" in all of its OP-$Q^2$I scores [23], [24]. In general, process is easier to measure, outcome is more important. Based on past related works [23], [24], the selected outcome and process $Q^2$Is are the following. (i) Outcome $Q^2$I: Effectiveness of the superpixel detection, estimated by means of two $Q^2$Is to be jointly maximized. (a) A VQ error, e.g., estimated as an RMSE, see Eq. (1), to be minimized. Vice versa, the inverse of the RMSE is a $Q^2$I to be maximized. (b) The number of image-objects, to be minimized by the image segmentation algorithm [90]. It is inversely related to a data compression rate, to be maximized. When the VQ error is zero, which means best case, because the number of segments is maximum and equal to the number of pixels, then segmentation is worst case because there is no pixel aggregation at all. An optimal compromise between these two mutually opposing $Q^2Is$ of effectiveness should be searched for. (ii) Process $Q^2I$: Efficiency, specifically: (a) computation time, required to be linear in the image size, and (b) central memory occupation, required to be kept "low" to cope with massive images. (iii) Process $Q^2I$: Degree of automation (ease of use), monotonically decreasing with the number of system's free-parameters to be user-defined. Full automation is required (refer to Section I). (iv) Process $Q^2I$: Robustness to changes in the input dataset. Data-independence is required (refer to Section I). (v) Process $Q^2I$: Robustness to changes in input parameters, if any. (vi) Process $Q^2I$: Scalability, to keep up with changes in users' needs and sensor properties. (vii) Outcome $Q^2I$: Timeliness, defined as the time between data acquisition and data-derived information product generation, to be minimized. (viii) Outcome $Q^2I$: Costs, in (a) manpower and (b) computer power, to be minimized. It is noteworthy that in papers published in the CV and RS literature, a CV system $Q^2A$ policy typically estimates the product effectiveness and, in case, the process efficiency, although these $Q^2Is$ are *per se* insufficient to assess the system's overall degree of operativeness.

### *C. Comparison with Alternative Approaches*

The proposed RGB image QNQ transform for superpixel detection and $Q^2A$ fully complies with the Marr definition of raw primal sketch [4], refer to Section II.A. As a consequence, it is preliminary not alternative to any inductive semi-automatic and site-specific algorithm [35] for either: (a) texture segmentation (perceptual grouping) [1], [2], [13], equivalent to the Marr's low-level vision full primal sketch [4], or (b) surface segmentation, equivalent to the Marr's high-level 2½D sketch [4], such as that presented in [42].

In addition, the proposed deductive automatic and data-independent superpixel detector should not be compared against any inductive semi-automatic and site-specific superpixel detection algorithm, such as the SLIC reviewed in Section II.F [41]. Belonging to two complementary rather than alternative families of inference algorithms for VQ, their process and outcome $Q^2Is$ are expected to feature complementary, rather than alternative behaviors, refer to Section II.D. This theoretical expectation is verified below.

(i) The RGB image QNQ transform realization detects each superpixel as a connected set of pixels featuring the same color label in a multi-level color map, whose map legend is a color dictionary defined *a priori*. Provided with a segment ID together with a color label, this superpixel is a planar objects identified by the Geospatial Consortium (OGC) terminology as 0D Point (code 1), 1D LineString (code 2), 2D Polygon (code 3), MultiPoint (code 4), MultiLineString (code 5), MultiPolygon (code 6) [45]. Vice versa, superpixels without a color label, such as those detected by the SLIC, can be coded as an OGC's Point, LineString or Polygon, on theory. In practice, the SLIC requires superpixels to be equally-sized and compact, which means their OGC spatial type is expected to be Polygon (code 3) exclusively. A color label provides each image-object with a semi-symbolic information superior to that of sub-symbolic pixels, whose semantics is zero, but never superior to, i.e., always less than or equal to, semantics of target 4D objects, e.g., land cover classes depicted in an EO image. The automatic detection in near real-time of semi-symbolic image-objects agrees with the quote by Marr that "vision goes symbolic almost immediately, right at the level of zero-crossing (first-stage primal sketch), without loss of information" ([4]; p. 343), refer to Section II.A.

(ii) In the implemented RGB image QNQ transform the shape of a superpixel can be any, e.g., an elongated superpixel, coded as LineString (code 2), features low compactness. In the SLIC algorithm, compact superpixels are parameterized by a compactness index $m$ to be user-defined in range [1, 40].

(iii) In the implemented RGB image QNQ transform the superpixel size can be any, from a minimum superpixel area value equal to 1 up to a maximum superpixel area equal to the image size $N$. As a consequence, their total number can be any, from 1, i.e., there is one single superpixel across the whole image, up to $N$, i.e., there is one superpixel per pixel. In the SLIC algorithm, the number $k$ of superpixels to be detected must be user-defined in advance and superpixels are forced to be equally-sized.

The conclusion is that there would be neither conceptual nor practical rationale in the direct comparison of process $Q^2$Is and outcome $Q^2$Is collected from the proposed deductive automatic superpixel detector and an inductive semi-automatic superpixel detection algorithm, such as the SLIC proposed in [41]. In practice, this comparison would be as pointless as comparing an inductive inference system with its initial conditions, to be provided by deductive inference, refer to Section I.

## IV. Materials

Input RGB images were selected to test the RGB image QNQ transform implemented as a proof of concept. This proof-of-concept was required to comply with process and outcome requirements, e.g., data-independence and robustness to changes in the input dataset, provided with metrological/statistically-based $Q^2$Is, refer to Section III.B. Therefore, three uncalibrated RGB images acquired by different imaging sensors mounted on different acquisition platforms, specifically, terrestrial, airborne and spaceborne, were selected for testing purposes.

Although the implemented RGB image QNQ transform was conceived for RS applications, it applies to any RGB image, e.g., an image depicting an "environmental scene" or an "object view". By definition [100], an "object view" subtends 1 to 2 meters around the observer, a "view on a scene" begins when there is actually a larger space between the observer and the fixated point, usually after 5 meters. In a real-world "object view", the observer's familiarity with the depicted object works as "ground truth", i.e., the quality of the RGB image QNQ transform can be intuitively assessed visually. For this reason and to prove the claim that the RGB image QNQ transform applies to any RGB image, one RGB jpeg-format image of a human face acquired by a consumer-level Canon color camera (of course, provided with no radiometric calibration capability), featuring rows = $RW$ = 230 and columns = $CL$ = 219, acquired by a consumer-level color camera, was collected by means of a web-based search-by-noun engine, see Fig. 8. One non-calibrated false-color (R channel = Visible Red, G channel = Near Infrared, B channel = Visible Blue) three-band image of a public event in Munich, Germany, acquired in 2013 by a 700 m-high airborne platform with a 4-band LEICA Airborne Digital Scanner (ADS) 80, with $RW$ = 2744 and $CL$ = 4616, was provided by the German Aerospace Center (DLR), see Fig. 9. One heterogeneous bi-temporal RGB synthetic aperture radar (SAR) image, with $RW$ = 4480 and $CL$ = 5012, acquired by the COSMO-SkyMed SAR sensor, was provided by the University of Naples Federico II, Italy [91]. Such an RGB-SAR image consists of the following bi-temporal heterogeneous SAR variables. R channel: interferometric coherence value in range [0, 1] coded as float; to save memory space by a factor of 4:1, it can be byte-coded into range {0, 255} with a negligible discretization error, equal to ((1./255)/2.) = 0.2% (where factor 2 in the denominator is due to rounding the ratio 1./255 to the closer integer, either superior or inferior). G channel: Test image at time T1 (e.g., rain season), SAR backscatter in range [0, 1], byte-coded into range {0, 255}. B channel: Reference image at time T0 < T1, SAR backscatter in range [0, 1], byte-coded into range {0, 255}, see Fig. 10.

## V. Results

From a methodological standpoint (refer to Section I and Section II.D), the goal of this experimental session was threefold. (a) Prove that the RGB image QNQ transform realization, where the software pipeline shown in Fig. 4 adopts the RGBIAM expert system as block 2 for VQ, is as a proof-of-concept able to meet the project requirements specified in Section I. Hereafter, it is referred to as the RGBIAM-QNQ realization.

(b) When different VQ approaches are employed as block 2 in Fig. 4 (refer to Section II.D and Section III.C), prove that: (i) deductive VQ, inherently automatic and non-adaptive to data, in comparison with inductive VQ, inherently semi-automatic and site-specific, feature complementary, rather than alternative functional properties. (ii) A hybrid VQ system, where deduction initializes induction (refer to Section I and Section II.D), overcomes limitations of each individual part.

To accomplish these experimental objectives, three VQ approaches were tested as block 2 in Fig. 4.

(I) A one-pass deductive RGBIAM discretization prototype with 49+1 quantization levels, including category "unknown", actually empty.

(II) A traditional iterative unlabeled data learning *k*-means VQ algorithm, implemented in the ENVI commercial software toolbox [62], with free-parameters $k$ = 49, max number of iterations $I$ = 3, change threshold = 5%, with centroids initialized by random sampling of the training dataset (refer to Section II.D). In agreement with an *n*-fold cross-validation approach adopted in [54], a three-fold cross-validation was applied to the *k*-means VQ algorithm. It was trained in each of the three test images, where the training error was assessed, refer to cells depicted in dark gray in Table 1 and Table 2. After reaching convergence, it was run upon the remaining two images, to assess its prediction error. The final VQ error was estimated as the sum of training and testing errors. To guarantee a fair accuracy assessment of the *k*-mean VQ algorithm adopted as block 2 in the software pipeline shown in Fig. 4, its VQ error was not estimated at the output of block 2, where each pixel is assigned to the closest center of mass in a codebook of $k$ global (image-wide) data-centroids, but at the output of block 6, where a piecewise-constant approximation of the input RGB image was synthesized. Dealing with local rather than global statistics, the latter VQ error is never superior, i.e., it is always inferior or equal, to the former.

(III) The RGBIAM's output 49-level color map of each test image was employed to initialize the $k$ = 49 data-centroids of the *k*-means VQ algorithm, with free-parameters to be user-defined equal to: max number of iterations $I$ = 3, change threshold = 5%.

Table 1 and Table 2 show multiple heterogeneous OP-$Q^2$Is, but computation time (refer to Section III.B), collected from the three test algorithms ran upon each of the three test RGB images (refer to Section IV). Output products are shown in Fig. 8 to Fig. 10. To accomplish a multivariate analysis of a heterogeneous set of OP-$Q^2$Is, featuring different unit of measure, range of change and sensitivity to changes in the input dataset, each univariate quantitative variable was z-scored (standardized) to feature zero mean and unit variance across test datasets per algorithm. Next, univariate non-dimensional z-scores were summed into a univariate "ultimate" score per algorithm, to allow inter-algorithm score comparisons, refer to Table 2. Hence, conclusions about inter-algorithm comparisons of process $Q^2$Is and outcome $Q^2$Is should stem from Table 2, generated from Table 1 by z-scoring.

In Table 3 the measured computation time of the RGBIAM-QNQ transform realization is shown as a dependent variable of the independent image size for each of the three input images. These time values include both a data processing time, relying on a fast central memory, and an I/O data time, involved with a slow secondary memory, when the software pipeline's output products include two color maps, two segmentation maps, four contour maps and two piecewise-constant RGB image approximations, refer to Section III.A. Because the data processing time is estimated to be linear in the image size (refer to Section III.A), the I/O data time is expected to increasingly dominate the data processing time when the image size increases.

Additional examples where the proposed RGBIAM-QNQ transform is employed for automatic analysis and regionalization of various 3-tuple combinations of either homogeneous or heterogeneous 2-D gridded variables are shown in Fig. 11 and Fig. 12 respectively.

## VI. Discussion

Introduced in Section V, test products, shown in Fig. 8 to Fig. 10, process $Q^2$Is and outcome (product) $Q^2$Is, collected in Table 1 to Table 3, are discussed below.

(i) Outcome $Q^2$I: Effectiveness, where two $Q^2$Is (vice versa, cost variables) must be jointly maximized (vice versa, minimized), specifically: (a) the RMSE and (b) the number of segments or, vice versa, the mean image-object area, inversely related to the number of segments. See Fig. 8 to Fig. 10 for perceptual assessment of the three different VQ approaches adopted in block 2 of the software pipeline shown in Fig. 4. In Table 2, generated from Table 1 by z-scoring, best results are shown in bold in the bottom row. Noteworthy, they slightly differ from those shown in bold in Table 1. In Table 2, in perfect agreement with theoretical expectations (refer to Section I), first, the mean image-object area is maximized by the data-independent RGBIAM expert system for VQ. Second, the RMSE is minimized across the three input datasets by the hybrid VQ approach, where the RGBIAM expert system initializes the learning-from-data *k*-means algorithm for VQ as block 2 in the software pipeline shown in Fig. 2.

(ii) Process $Q^2$I: Efficiency of the RGBIAM-QNQ realization in terms of: (a) memory occupation and (b) computation time. The RGBIAM-QNQ software pipeline is implemented in tile streaming mode, suitable for massive data mapping problems, where the maximum dynamic memory occupation is fixed, irrespective of the image size. In these experiments the dynamic memory maximum size parameter was set equal to 800 MB of random access memory (RAM), which can be considered a “small” value for dynamic memory occupation in standard personal computers. In addition to these 800 MB of RAM, an estimate of the dynamic memory occupation required by the STD was provided in Section III.A. About computation time, the computational complexity of the RGBIAM-QNQ realization was claimed to be linear in the image size (refer to Section III.A). Therefore, data processing time was expected to be linear in the image size. When it was run on a Dell Power Edge 710 server with dual Intel Xeon @ 2.70 GHz processor with 64 GB of RAM and a 64-bit Linux operating system, the RGBIAM-QNQ realization required a computation time, including data processing and data I/O operations (refer to Section V), shown in Table 3. It agrees well with the linear-time expectation. The system output rate, defined as the inverse of computation time, can be considered not inferior to a reasonable input rate. For example, the typical input rate of a massive EO image acquired by a geostationary spaceborne EO imaging sensor is one every 15 minutes. If this consideration holds true, then computation time of the proposed RGBIAM-QNQ realization can be considered near real-time.

(iii) Process $Q^2$I: Degree of automation of the RGBIAM-QNQ realization. It requires neither user-defined parameter nor training dataset to run, i.e., it is fully automatic. Hence, its degree of supervision is zero. Vice versa, its degree of automation is maximum and cannot be surpassed by any alternative approach.

(iv) Process $Q^2$I: Robustness to changes in the input dataset of the RGBIAM-QNQ realization. It is non-adaptive to input data, hence its robustness to changes in the input dataset is maximum and cannot be surpassed by alternative approaches.

(v) Process $Q^2$I: Robustness to changes in input parameters of the RGBIAM-QNQ realization. It requires no user-defined parameter to run. Hence, its robustness to changes in input parameters is maximum and cannot be surpassed by alternative approaches.

(vi) Process $Q^2I$: Maintainability/ scalability/ re-usability, to keep up with changes in users' needs and sensor properties. The proposed RGBIAM-QNQ realization can be applied to any existing or future planned RGB imaging sensor, including consumer-level color cameras mounted in smartphones or on board light-weight UAVs, whether or not radiometrically calibrated, in true colors or false colors (e.g., refer to Section IV), irrespective of the image size. Rather than as a standalone low-level vision subsystem, it should be employed at the first stage of an innovative hybrid feedback IUS architecture [23], [24], consisting of six stages, including stage 0 (zero) for image pre-processing, suitable for OBIA applications whose final output product consists of symbolic image-objects in a GIS-ready file format. Shown in Fig. 13, this novel IUS design is completely alternative to the mainstream inductive feedforward IUS architecture adopted by the large majority of the RS and CV communities, e.g., support vector machines [36], random decision forest classifiers [37] and deep convolutional image neural networks [38], [39] (refer to Section I).

(vii) Outcome $Q^2I$: Timeliness, defined as the time span between data acquisition and product generation. The proposed RGBIAM-QNQ realization reduces timeliness from image acquisition to information product generation to almost zero, due to linear-time computation exclusively, because user's supervision is zero. On the contrary, timeliness of inductive semi-automatic and site-specific data learning algorithms, such as the inductive k-means VQ algorithm, is always superior to zero, due to the time spent by a human supervisor in selecting and collecting a training dataset representative of the complexity of the inductive learning-from-data problem, in addition to the computation time spent by the algorithm to learn-from-data before reaching convergence (stability) at the cost of plasticity.

(viii) Outcome $Q^2I$: Costs, monotonically increasing with manpower and computer power. The RGBIAM-QNQ realization is prior knowledge-based and near real-time in a standard laptop computer. Its costs are zero in terms of user's supervision and almost negligible in terms of computational power. This does not hold for inherently semi-automatic and site-specific inductive data learning algorithms, such as the inductive *k*-means VQ algorithm. They require human supervision to define the system's free-parameters and to provide a testing dataset. It is worth mentioning that many existing real-time solutions in image and video segmentation are based on multi-scale window-based local statistic estimates. These algorithms are computationally very intensive on traditional central processing units (CPUs). In practice, they require more expensive information technology solutions, with parallel hardware to estimate subwindows independently and graphics processor units (GPUs) for acceleration purposes [88].

(ix) Process $Q^2I$: Data compression rate, equal to 4:1 up to 6: 1, depending on the selected number of quantization regions, equivalent to color names.

According to the definition provided in Section III.B, since it scores "high" in each of the selected OP-$Q^2Is$, the RGBIAM-QNQ realization can be considered off-the-shelf, i.e., ready-for-use. In particular, it is: (a) eligible for inclusion in a CV software library of off-the-shelf low-level vision functions, such as OpenCV [92], and (b) suitable for use in mobile software applications, defined as lightweight computer programs specifically designed to run on web browsers and mobile devices, such as tablet computers and smartphones.

## VII. Conclusions

Outpaced by the rate of collection of images and videos of ever-increasing quality and quantity, such as those acquired by consumer- or commercial-level color cameras mounted in mobile devices or on unmanned aerial vehicles (UAVs), the CV and remote sensing (RS) communities are affected by an ongoing lack of *information-as-data-interpretation* capabilities [21]. To invert this trend, the present research and technological development (RTD) CV software project promotes the exploitation of *a priori* knowledge, available in addition to sensory data, to initialize inductive learning-from-data algorithms in agreement with biological cognitive systems, where genotype initializes phenotype. A multidisciplinary cognitive approach to vision (see Fig. 1), where image pre-processing and understanding are considered *hybrid* (combined deductive and inductive) inference problems, is in contrast with the mainstream RTD in computer vision and RS, focused on feedforward inductive image learning systems, such as support vector machines [36], random decision forest classifiers [37] and deep convolutional image neural networks [38], [39]. It is well known, but rather oversighted by the scientific community that inductive learning-from-data is an inherently ill-posed inference problem, which requires *a priori* knowledge in addition to data to become better posed for numerical treatment [32], [33].

In compliance with the Quality Assurance Framework for Earth Observation (QA4EO) requirements, the goal of the present low-level (pre-attentional) vision software design and implementation project was to develop an off-the-shelf hybrid reversible quantitative-to-nominal-to-quantitative (QNQ) transform of a monitor-typical red-green-blue (RGB) image for automatic superpixel detection and quantitative quality assurance ($Q^2$A) in linear time. Any stack of three scalar 2-D gridded variables, for example, three heterogeneous geospatial variables such as a vegetation spectral index, a geospatial density function of population and a geospatial index of scholarization, can be selected by a user to be investigated as a monitor-typical RGB image. Whenever it is transformed in a monitor-typical RGB image, any three-channel combination of 2D gridded variables can be automatically investigated by the proposed expert system for RGB image QNQ transform. Irrespective of its implementation the proposed six-stage system design, shown in Fig. 4, simultaneously fills two traditional information gaps: from pixels to image-objects and from ever-varying sensory data to stable concepts, such as data-, user- and application-independent color names belonging to a color dictionary community-agreed upon in advance, equivalent to *a priori* visual knowledge. In the direct data coding phase, the input RGB image is automatically partitioned into connected image-objects provided with a color label in linear time. These semi-symbolic planar entities feature some degree of semantics, e.g., color green is typical of vegetation. Known in the existing literature as texture elements, texels, tokens or superpixels, color-homogeneous planar entities belong to the raw primal sketch of the Marr's low-level vision system model [4]. In the inverse data decoding phase, a superpixelwise-constant RGB image reconstruction provides each pixel with an approximation error. By improving the structural and procedural knowledge of the static decision tree implemented as the deductive vector quantization (VQ) block 2 of the RGB image QNQ transform pipeline shown in Fig. 4, pixel-based approximation errors can be maintained below the target visual problem's VQ error requirement, e.g., see Fig. 8(t) to Fig. 8(v).

A realization of the low-level vision system design for automatic linear-time detection and $Q^2$A of semi-symbolic superpixels was presented as a proof-of-concept. (I) It complies with the quote by Marr that "vision goes symbolic almost immediately, right at the level of zero-crossing (first-stage primal sketch), without loss of information" ([4]; p. 343). (II) It is complementary not alternative to inductive ill-posed semi-automatic and site-specific data learning algorithms for either VQ, such as the *k*-means and ISODATA algorithms, or image segmentation algorithms related to the Marr's raw primal sketch, such as the SLIC [41], full primal sketch [13] or 2½ sketch [74] (refer to Section II.A). For example, adopted as an automatic near real-time edge-preserving image smoothing filter, e.g., see Fig. 8(p) to Fig.8(r), the proposed RGB image QNQ converter can enhance segmentation of a video sequence pursued by a Metropolis algorithm [88], [89]. (III) By scoring "high" in each of the selected outcome and process $Q^2$Is (OP-$Q^2$Is), it can be considered worth to enrich a general-purpose low-level vision software library, such as OpenCV [92], because presumably useful to a broad audience. For example, it is suitable for use in mobile software applications, defined as lightweight computer programs specifically designed to run on web browsers and mobile devices, such as tablet computers and smartphones. (IV) Rather than being considered a standalone low-level vision module, any *a priori* knowledge-based RGB image QNQ transform can be adopted in the low-level vision first stage of a novel hybrid feedback IUS architecture, shown in Fig. 11. It is well known that color names "cannot always be inverted to unique land cover class names" [93]. To disambiguate one-to-many or many-to-many relationships of color names with classes of target objects belonging to the 4D real world-through-time, spatio-temporal properties of image-objects must be investigated, in addition to first-stage color properties. A target object class-specific spatio-temporal analysis conditioned by first-stage color analysis can be performed at the high-level information processing stages 2 and 3 of the novel IUS architecture, shown in Fig. 11, according to a stratified convergence-of-evidence classification approach consistent with human reasoning [1], [2], [23], [24]. Hybrid inference mechanisms provide this novel IUS architecture with feedback loops, from higher to lower information processing stages. It means that low-level qualitative/categorical/nominal information products, such as pre-classification color maps, are not exclusively useful to better condition high-level classification tasks. They can also be adopted for statistical stratification of inherently ill-posed image pre-processing tasks, such as Earth observation (EO) image atmospheric correction and topographic correction, image co-registration, image compositing, etc. [94]. For example, the proposed RGB image QNQ converter can be applied to a bi-temporal RGB-SAR image, see Fig. 10, to enhance estimation of the coherence input variable for high-coherence targets, such as vessels [98], as well as to improve detection of urban areas described as high-texture image areas [1].

Planned future applications of the proposed off-the-shelf RGB image QNQ transform will try to provide both traditional inherently ill-posed low-level image enhancement algorithms, such as SAR image despeckling [95], [96], and high-level image understanding algorithms, such as SAR image classification [97], [98], with novel solutions based on *a priori* knowledge in addition to data. For example, an innovative low-level vision project called "Automatic speckle model-free despeckling of bitemporal RGB-SAR imagery" is currently ongoing [99].

## ACKNOWLEDGMENT

To accomplish this work Andrea Baraldi was supported in part by the National Aeronautics and Space Administration (NASA) under Grant No. NNX07AV19G, issued through the Earth Science Division of the Science Mission Directorate. Dirk Tiede was supported in part by the Austrian Research Promotion Agency (FFG), in the frame of project AutoSentinel2/3, ID 848009. Andrea Baraldi thanks Prof. Raphael Capurro for his hospitality, patience, politeness and open-mindedness. He also thanks Prof. Christopher Justice, Chair of the Department of Geographical Sciences, University of Maryland, for his friendship and support, together with Michael L. Humber, for his initial collaboration. The authors also wish to thank the Editor-in-Chief, Associate Editor and reviewers for their competence, patience and willingness to help.

## FIGURES AND FIGURE CAPTIONS

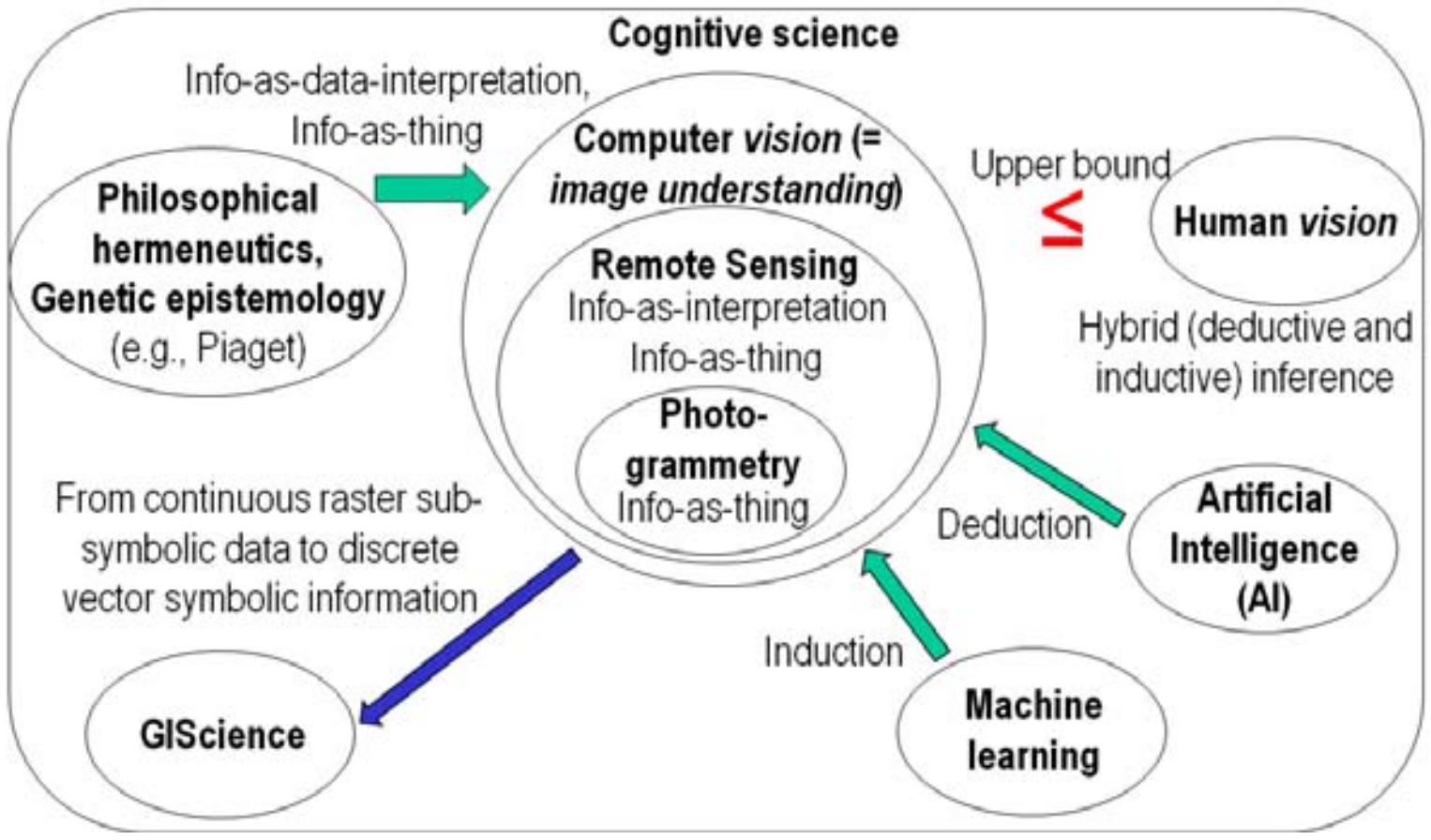

Fig. 1. Like engineering, remote sensing (RS) is a metascience, whose goal is to transform knowledge of the world, provided by other scientific disciplines, into useful user- and context-dependent solutions in the world. Cognitive science is the interdisciplinary scientific study of the mind and its processes. It examines what cognition (learning [28]) is, what it does and how it works. It especially focuses on how information/knowledge is represented, acquired from sensory data, processed and transferred within nervous systems (humans or other animals) and machines (e.g., computers) [26]-[28].

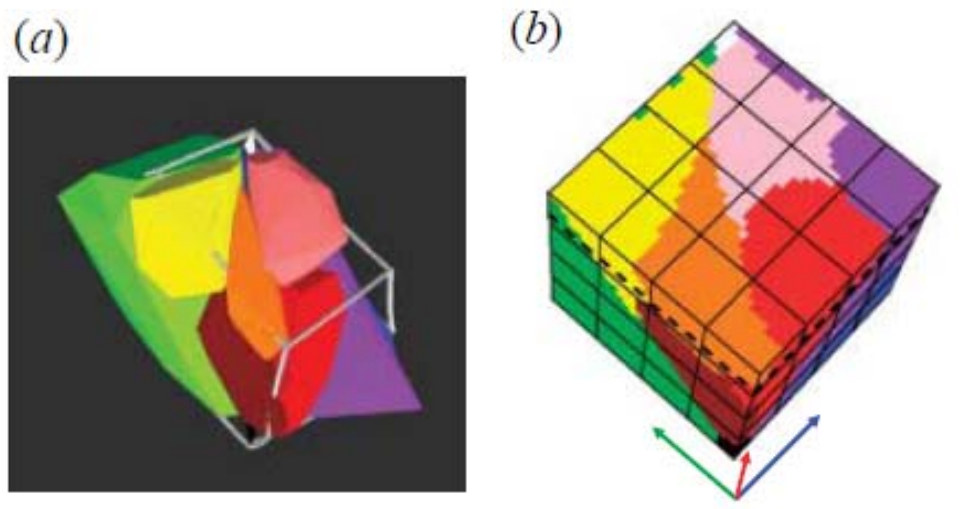

Fig. 2. Stages in mapping human basic colors (BCs) into the RGB cube. (a) CIE-Lab space with regions of eleven basic color labels as colored polyhedra and the edges of the monitor-typical RGB cube (in grey). (b) 323 quantization of the RGB space with the basic color extents from (a) mapped into it. The uniform 43 quantization of the RGB cube shown in (b) was adopted for representing color category systems whose classification performance was assessed. Images reproduced courtesy of [53].

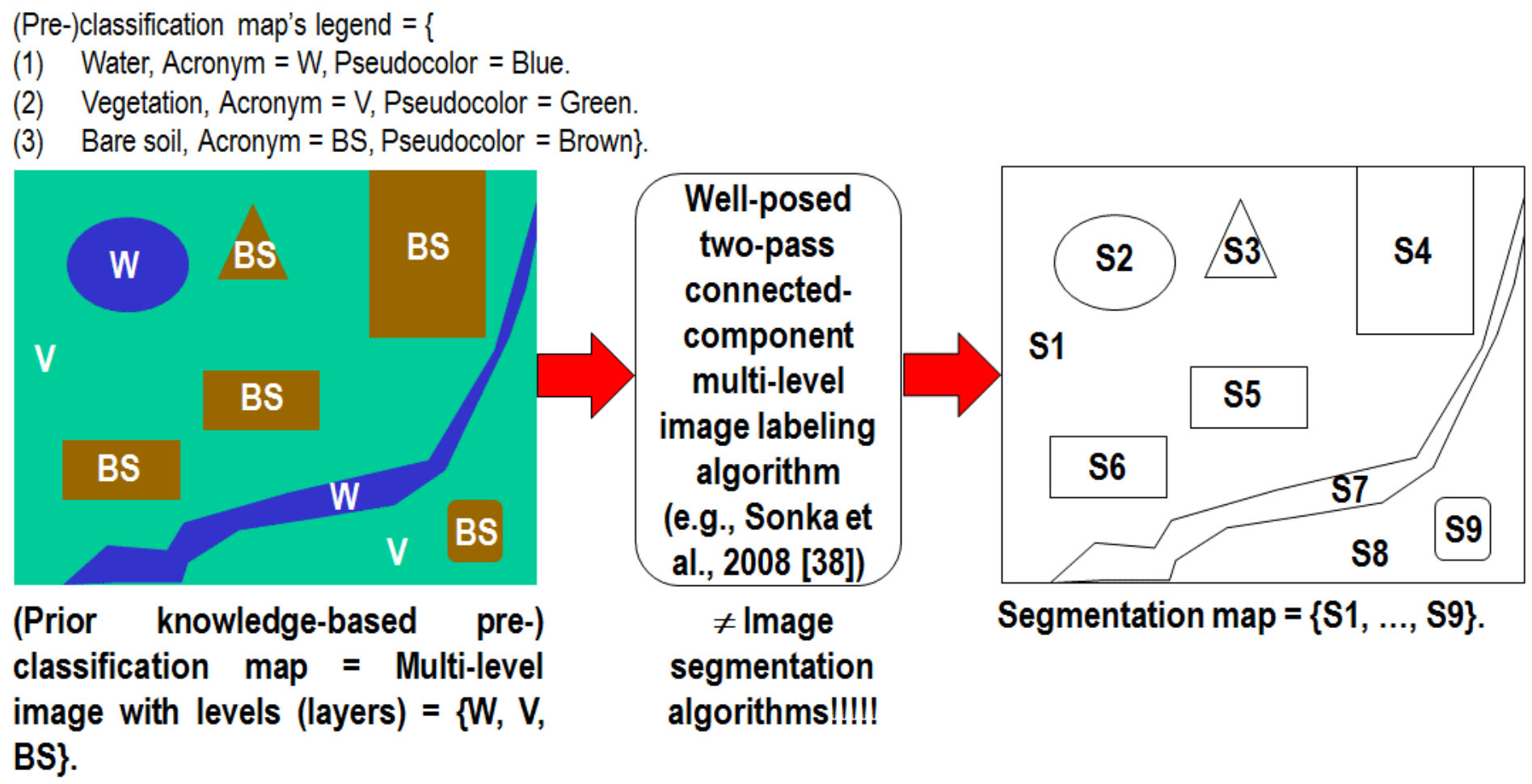

Fig. 3. One segmentation map is deterministically generated from one multi-level image, e.g., a classification map, but the vice versa does not hold, i.e., many multi-level images can generate the same segmentation map. In this example, stratum Vegetation, V, of the classification map consists of the two disjoint image objects, S1 and S9. Each image-object S1 to S9 consists of a connected set of pixels sharing the same label. Hence, the three spatial primitives labeled pixels, labeled segments and labeled strata (layers) co-exist in parent-child relationships.

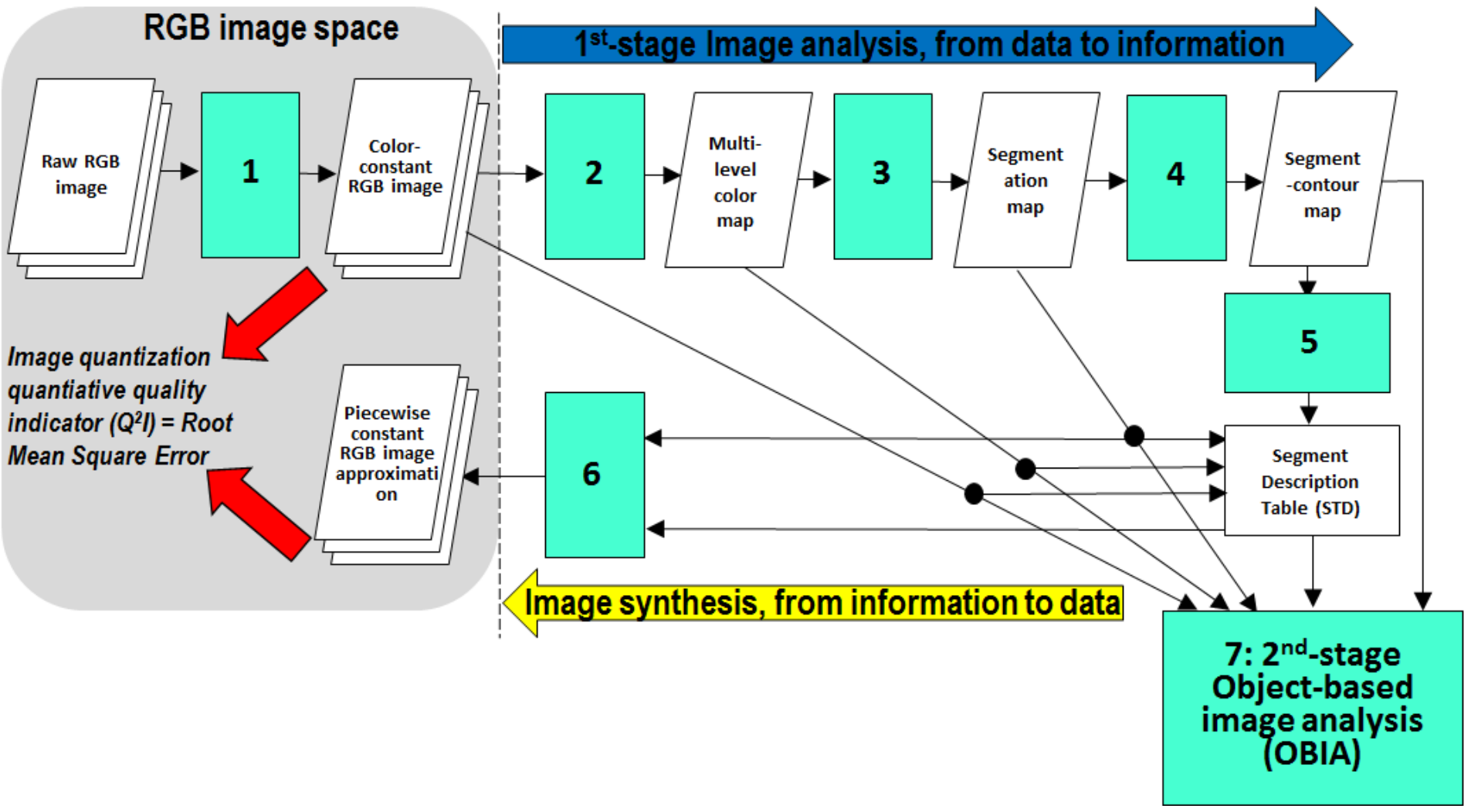

Fig. 4. The automated near real-time RGBIAM software toolbox for prior knowledge-based vector quantization, VQ, consists of subsystems 1 to 6. Phase 1-of-2 = Encoding phase/Image analysis - Stage 1: Self-organizing statistical algorithm for color constancy. Stage 2: Prior knowledge-based (static) RGBIAM decision tree for RGB cube partitioning (quantization, polyhedralization). Stage 3: Well-posed two-pass connected-component detection in the multi-level color map. Connected-components in the color map domain are connected sets of pixels featuring the same color label. These connected-components are also called image-objects, segments or superpixels. Stage 4: Well-posed superpixel-contour extraction. Stage 5: Well-posed Superpixel Description Table (STD) allocation and initialization. Phase 2-of-2 = Decoding phase/Image synthesis - Stage 6: Superpixelwise-constant input image approximation ("object-mean view") and per-pixel VQ error estimation. (Stage 7: in cascade to the RGBIAM's superpixel detection, a high-level object-based image analysis (OBIA) approach can be adopted).

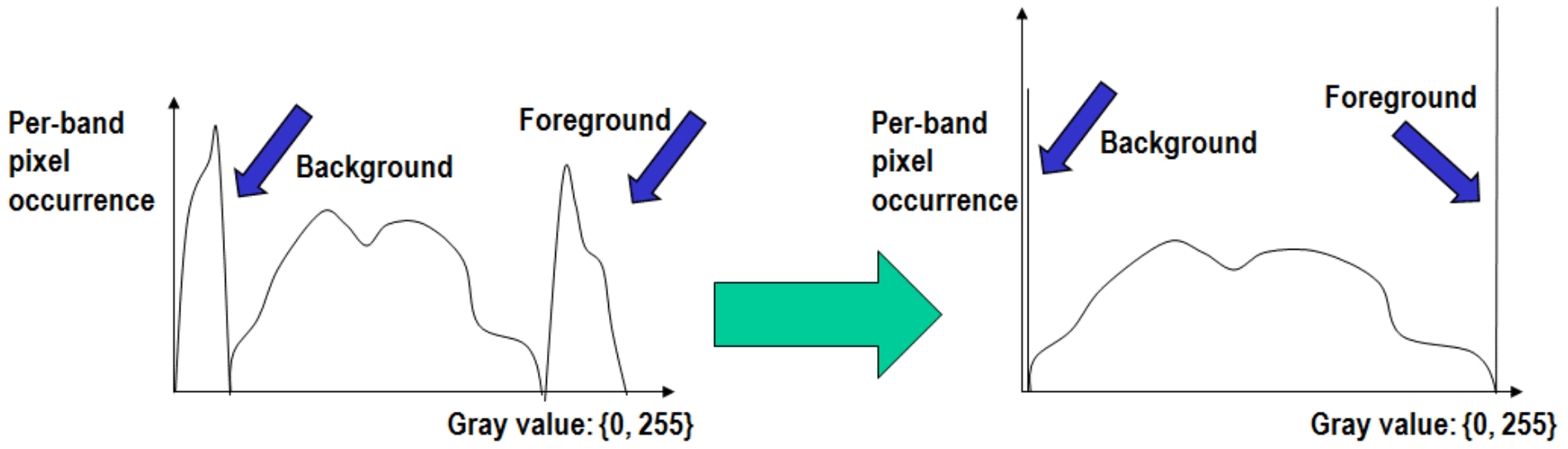

Fig. 5. Example of a gray-value univariate (one-channel) 1st-order distribution, to be stretched for color constancy. Four categories of univariate distributions can be considered: (i) neither a background nor a foreground mode is present in addition to a central mode; (ii) only a background mode with a right tail can be identified, (iii) only a foreground mode with a left tail can be identified, and (iv) both background and foreground modes are present in addition to a central mode.

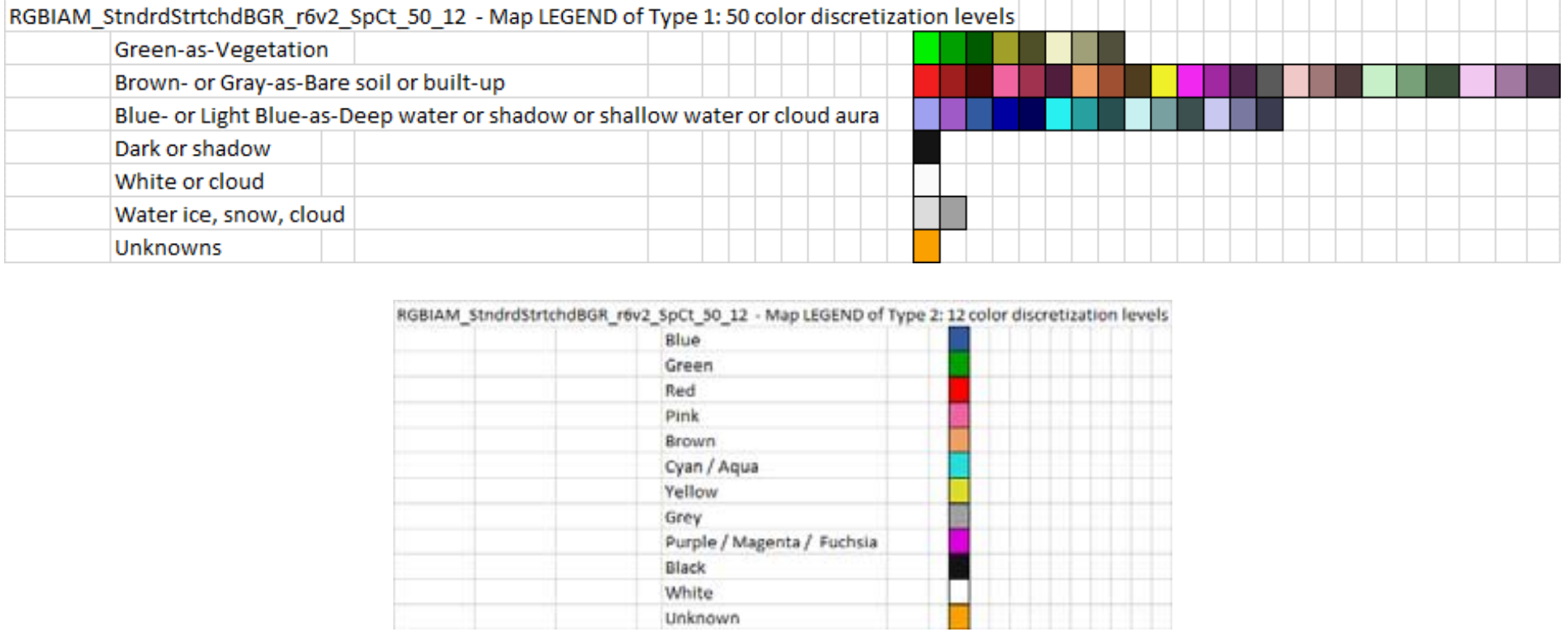

Fig. 6. Two-level RGB cube quantizer. Color map's legend at (left) fine (49 + 1 class unknown) and (right) coarse (11 + 1 class unknown) quantization levels, where the latter is a mutually exclusive and totally exhaustive combination of the former.

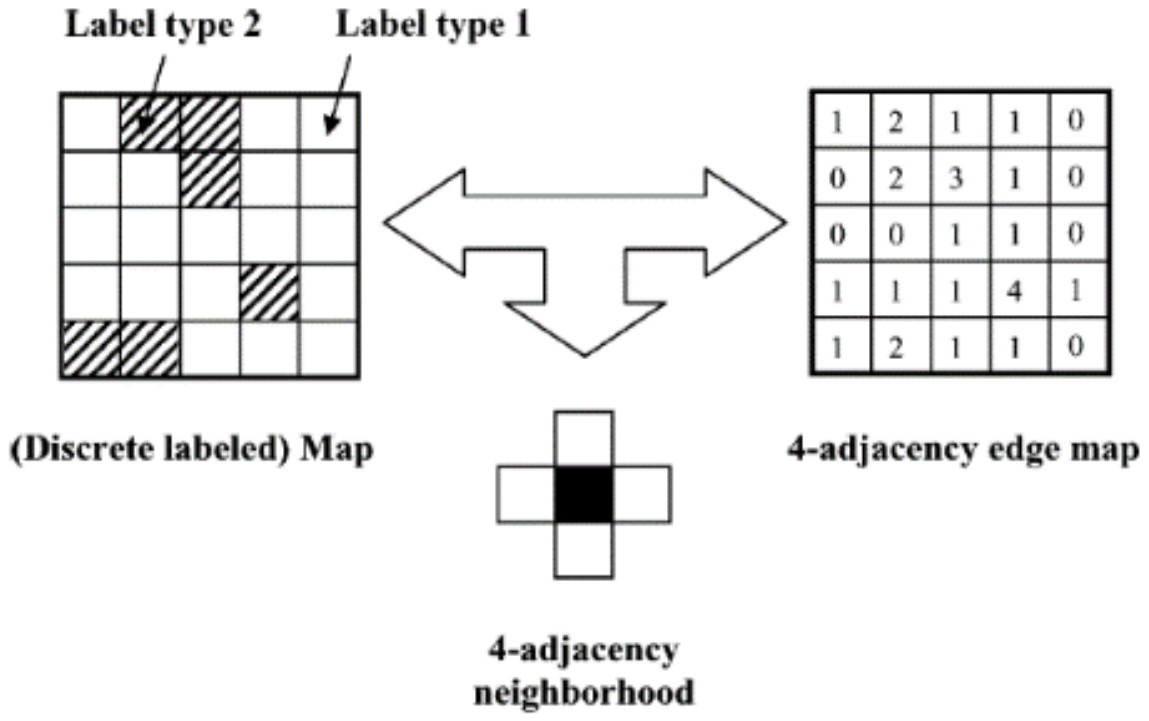

Fig. 7. Example of a 4-adjacency cross-aura map, shown at right, generated from a two-level image, shown at left [78], [85].

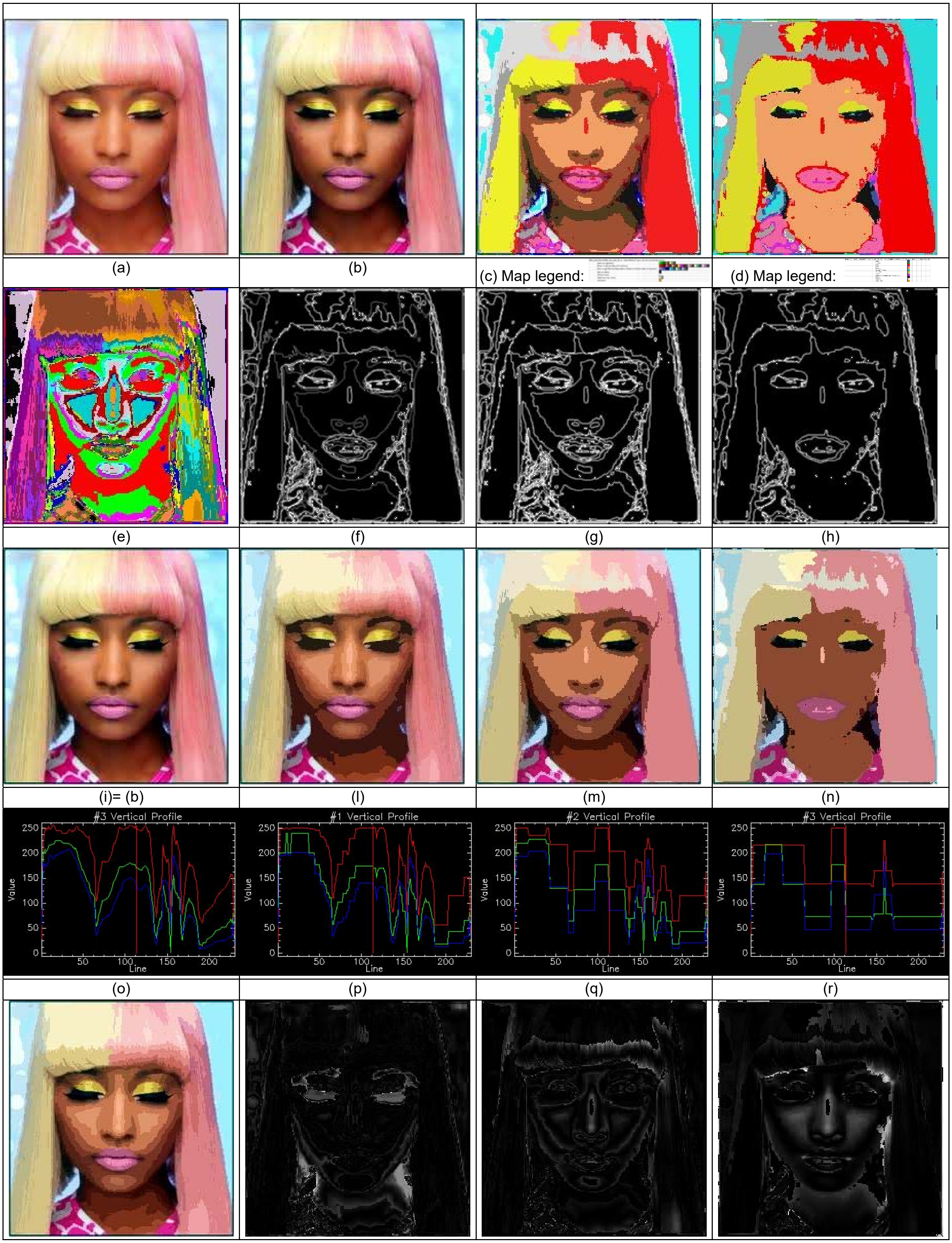

| | | | |
|---|---|---|---|
| | 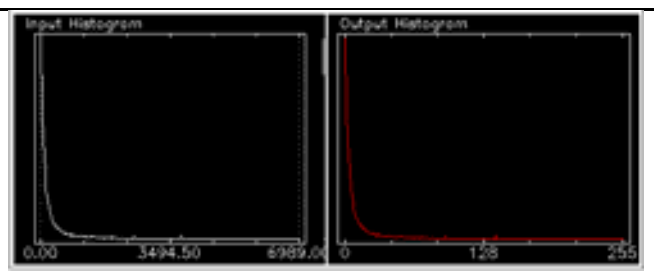 | 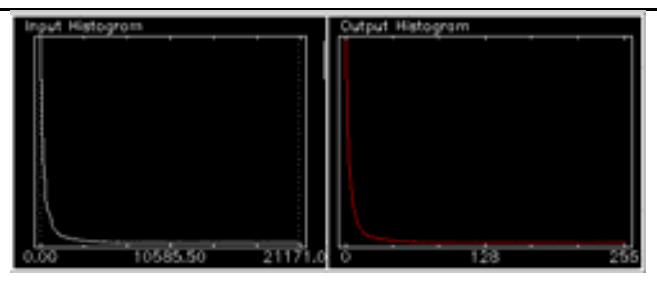 | 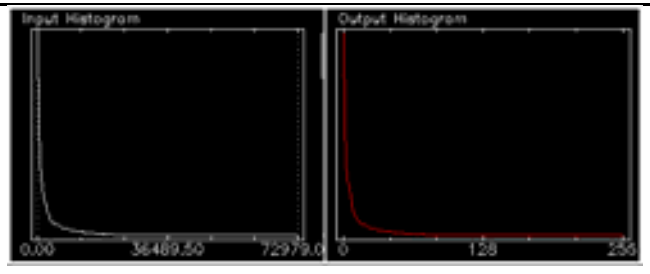 |
| | Square Error Sum, Basic Stats: Min = 0.0, Max = 6989.0, Mean = 306.877626, Stdev = 544.575772. | Square Error Sum, Basic Stats: Min = 0.0, Max = 21171.0, Mean = 803.150923, Stdev = 1230.231361. | Square Error Sum, Basic Stats: Min = 0.0, Max = 72979.0, Mean = 4145.427894, Stdev = 7967.770808. |
| (s) | (t) | (u) | (v) |

Fig. 8. Test image 1 of a human face. Image size: RW = 230, CL = 219. (a) Original RGB image acquired by a consumer-level color camera. (b) RGB image QNQ converter. Black 1, self-organizing color constancy. (c) RGBAIM-QNQ converter. Color map in pseudocolors, 49+1 color quantization levels. (d) RGBIAM-QNQ converter. Color map in pseudocolors, 11+1 color quantization levels. (e) *k*-means, free-parameters *k* = 49, Iterations = 3. Color map in pseudocolors. (f) Sum, in range {0, 16}, of the two 8-adjacency cross aura measures shown in (g) and (h). (g) 8-adjacency cross aura measure in range {0, 8}, generated from the RGBIAM-QNQ converter's color map, 49+1 color quantization levels. (h) 8-adjacency cross aura measure in range {0, 8}, generated from the RGBIAM-QNQ converter's color map, 11+1 color quantization levels. (i)= (b). (l) *k*-means, with *k* = 49, Iterations = 3. Object-based piecewise constant RGB image synthesis. (m) RGBIAM-QNQ converter. Object-based piecewise constant RGB image synthesis, 49+1 color quantizer. (n) RGBIAM-QNQ converter. Object-based piecewise constant RGB image synthesis, 11+1 color quantizer. (o)-(r) Vertical profile of image (i) to (n) respectively, located in the central column of the image. Edge-preserving smoothed image. (s) Hybrid VQ. Object-based piecewise constant RGB image synthesis. A first-stage deductive RGBIAM color partitioner automatically detects 49(+1 category "unknown, actually empty) centroids employed as input to a second-stage inductive *k*-means VQ, 3 iterations. (t)-(v) Per-pixel square quantization error sum across bands, when the reconstructed image (l) to (n), respectively, is compared with the input image (i) = (b).

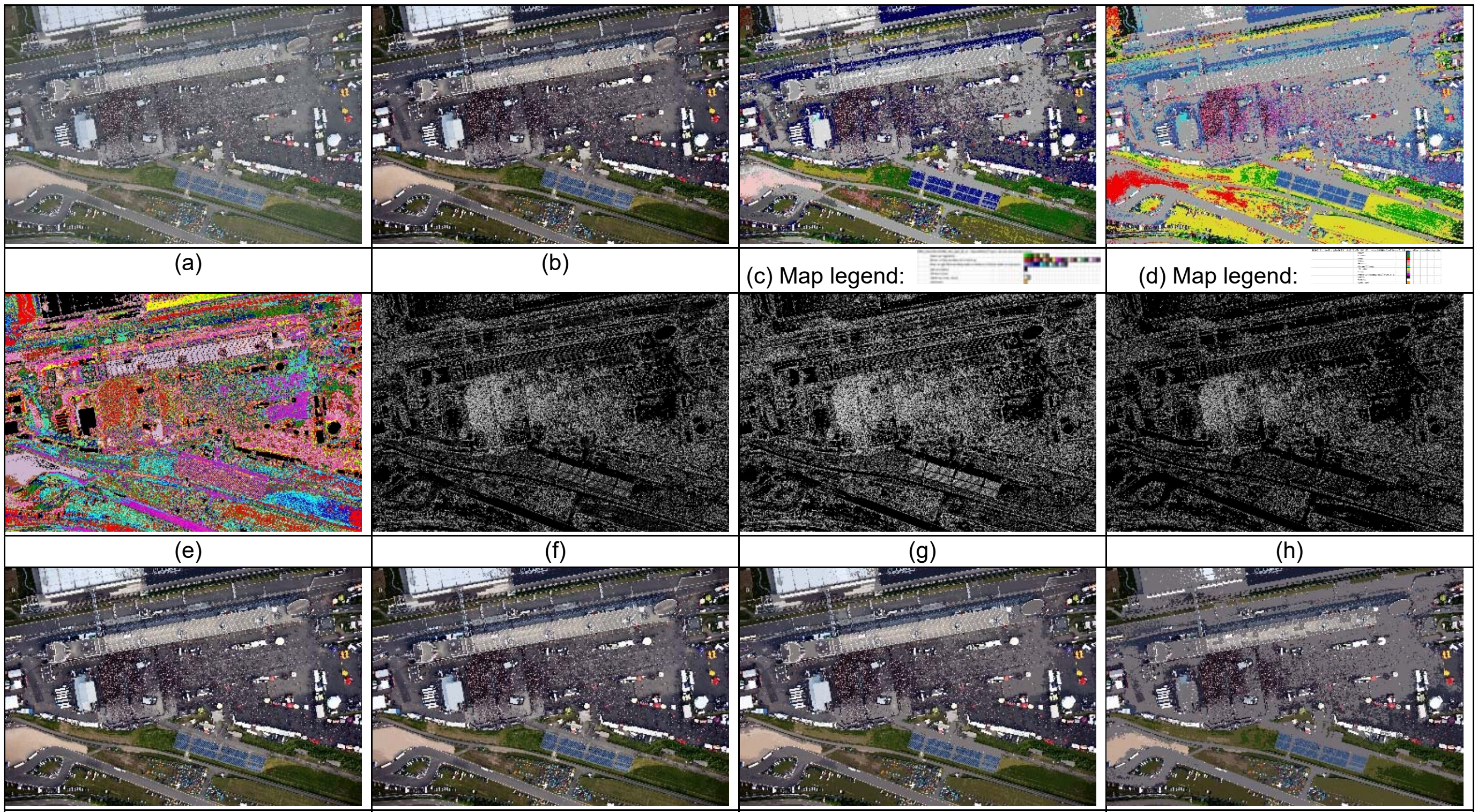

Fig. 9. Test image 2. 700 m-height aerial image of a public event in Munich, Germany (courtesy of DLR). Image size: RW = 2744, CL = 4616. (a) Original non-calibrated RGB image in false colors (R = Visible Red, G = Near Infrared, B = Visible Blue). (b) RGB image QNQ converter. Block 1, self-organizing color constancy. (c) RGBIAM-QNQ converter. Color map in pseudocolors, 49+1 color quantization levels. (d) RGBIAM-QNQ converter. Color map in pseudocolors, 11+1 color quantization levels. (e) *k*-means, free-parameters *k* = 49, Iterations = 3. Color map in pseudocolors. (f) Sum, in range {0, 16}, of the two 8-adjacency cross aura measures shown in (g) and (h). (g) 8-adjacency cross aura measure in range {0, 8}, generated from the RGBIAM-QNQ converter's color map, 49+1 color quantization levels. (h) 8-adjacency cross aura measure in range {0, 8}, generated from the RGBIAM-QNQ converter's color map, 11+1 color quantization levels. (i) = (b). (l) *k*-means, with *k* = 49, Iterations = 3. Object-based piecewise constant RGB image synthesis. (m) RGBIAM-QNQ converter. Object-based piecewise constant RGB image synthesis, 49+1 color quantizer. (n) RGBIAM-QNQ converter. Object-based piecewise constant RGB image synthesis, 11+1 color quantizer.

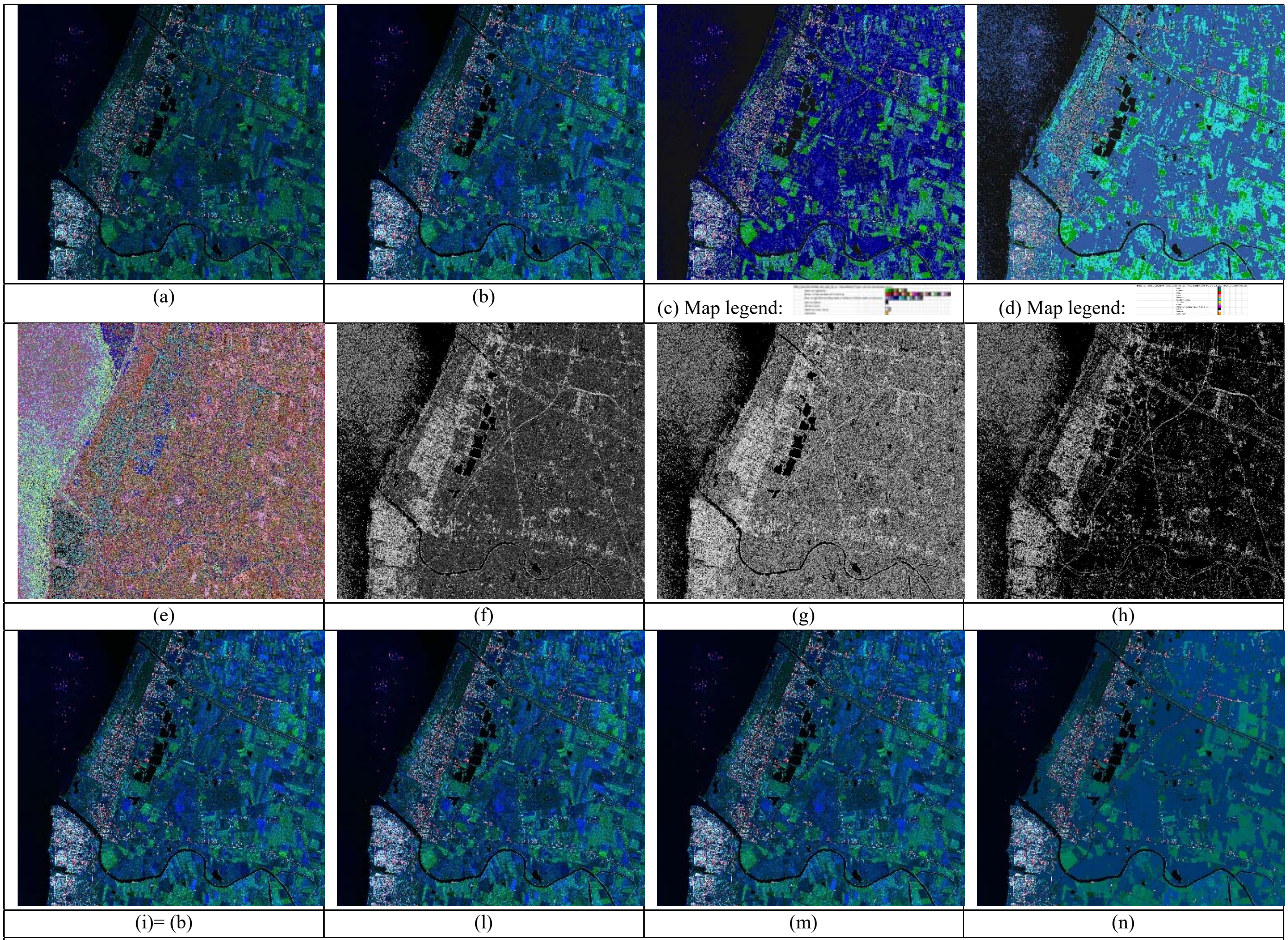

Fig. 10. Test image 3. Spaceborne bi-temporal RGB-SAR image of the Campania region, Italy (courtesy of University of Naples Federico II, Italy). Image size: RW = 4480, CL = 5012. (a) Original RGB-SAR image. (b) RGB image QNQ converter. Block 1, self-organizing color constancy. (c) RGBIAM-QNQ converter. Color map in pseudocolors, 49+1 color quantization levels. (d) RGBIAM-QNQ converter. Color map in pseudocolors, 11+1 color quantization levels. (e) *k*-means, free-parameters *k* = 49, Iterations = 3. Color map in pseudocolors. (f) Sum, in range {0, 16}, of the two 8-adjacency cross aura measures shown in (g) and (h). (g) 8-adjacency cross aura measure in range {0, 8}, generated from the RGBIAM-QNQ converter's color map, 49+1 color quantization levels. (h) 8-adjacency cross aura measure in range {0, 8}, generated from the RGBIAM-QNQ converter's color map, 11+1 color quantization levels. (i) = (b). (l) *k*-means, with *k* = 49, Iterations = 3. Object-based piecewise constant RGB image synthesis. (m) RGBIAM-QNQ converter. Object-based piecewise constant RGB image synthesis, 49+1 color quantizer. (n) RGBIAM-QNQ converter. Object-based piecewise constant RGB image synthesis, 11+1 color quantizer.

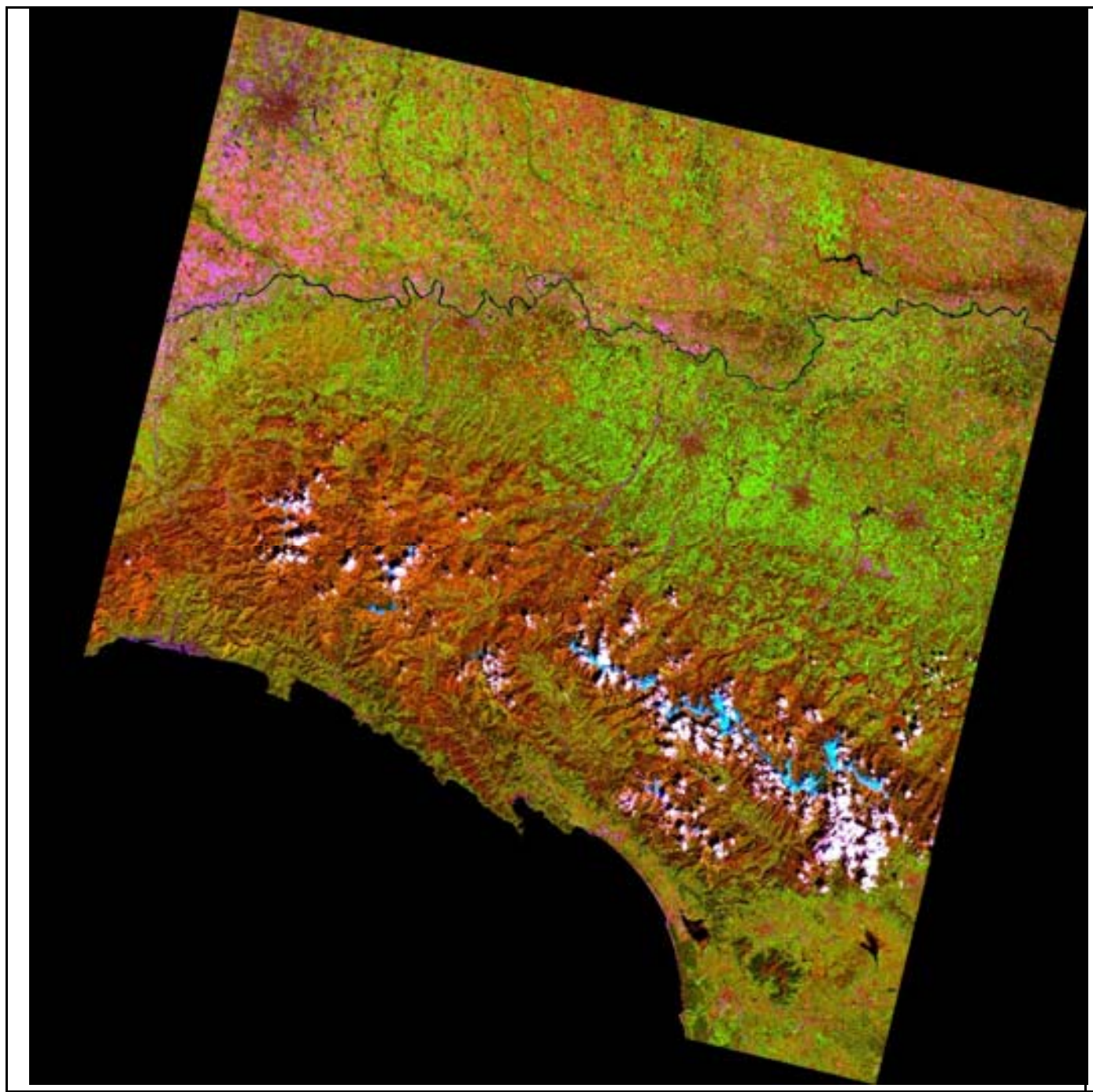

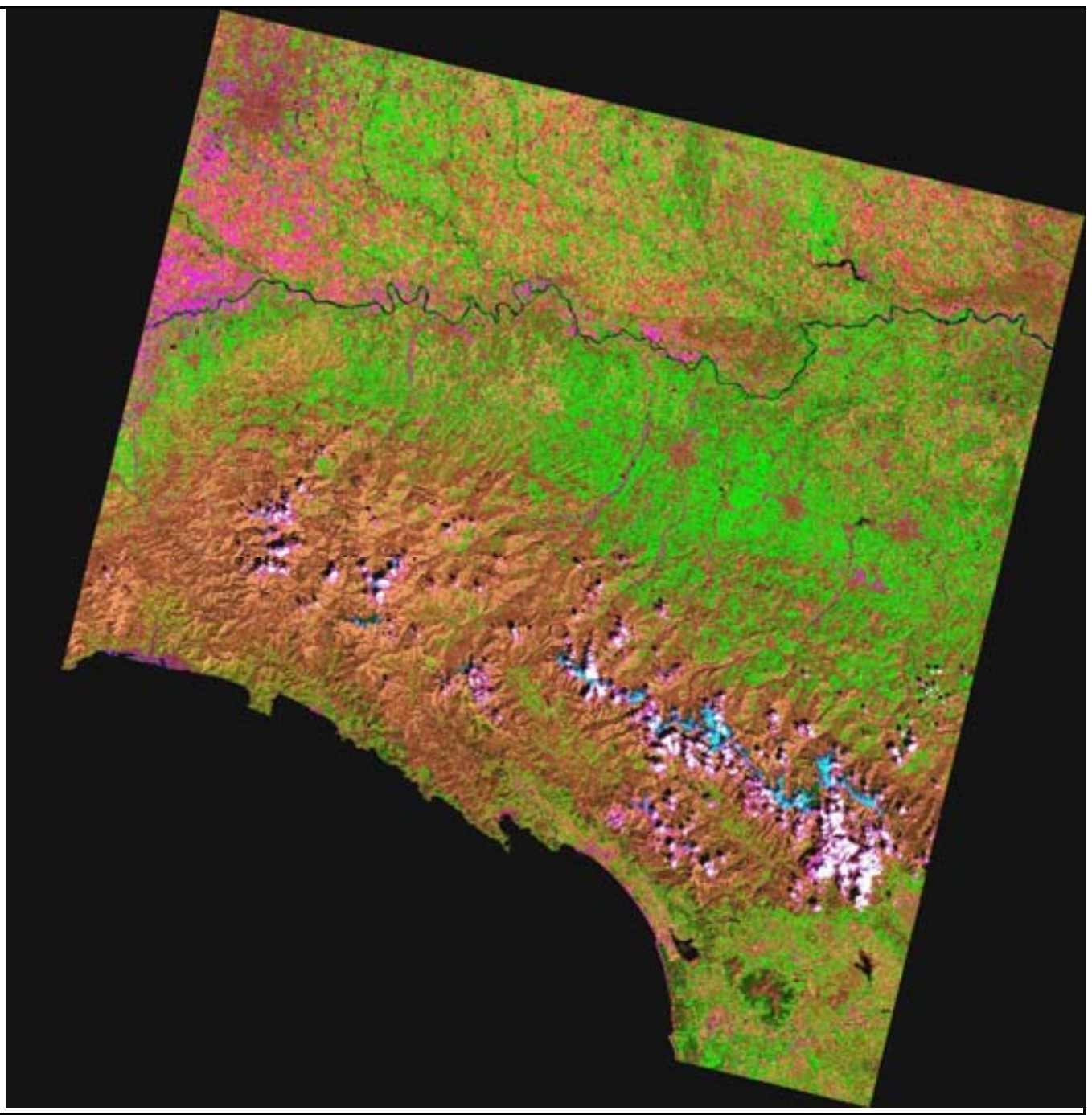

(a) RGB image histograms.

(b) Map histogram and legend.

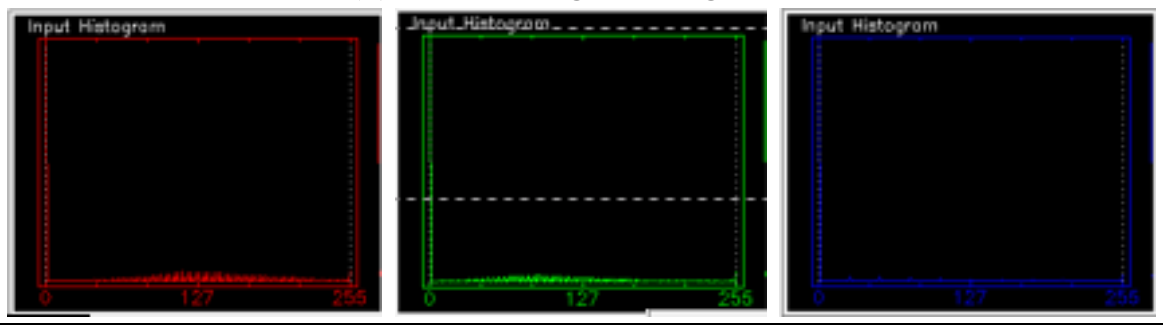

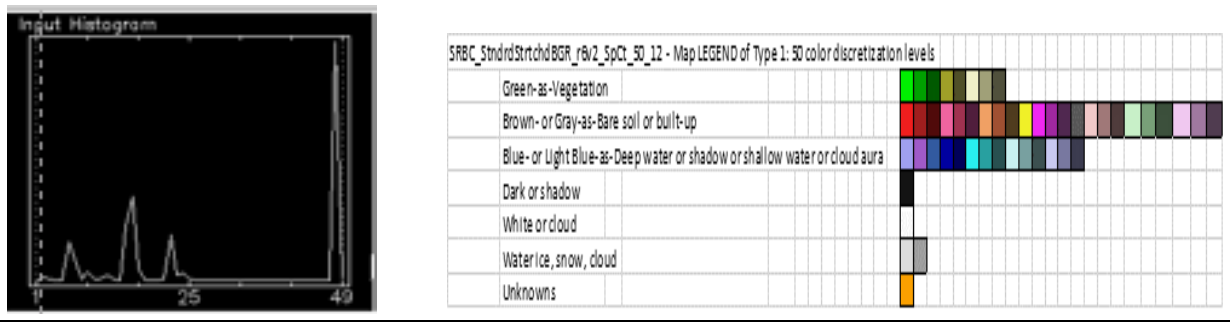

Fig. 11. (a) Test image 4. False-color RGB Landsat-8 OLI+TIRS image of Northern Italy, acquired on 2014-04-07, radiometrically calibrated into top-of-atmosphere reflectance values. Next, it was subject to a “default” histogram stretching for color constancy. In particular: R = Medium Infrared = Band 6, G = Near Infrared = Band 5, B = Visible Blue = Band 2. Spatial resolution: 30 m. No stretching is applied for visualization purposes.. (b) RGBIAM-QNQ converter. Color map in pseudocolors, 49+1 color quantization levels.

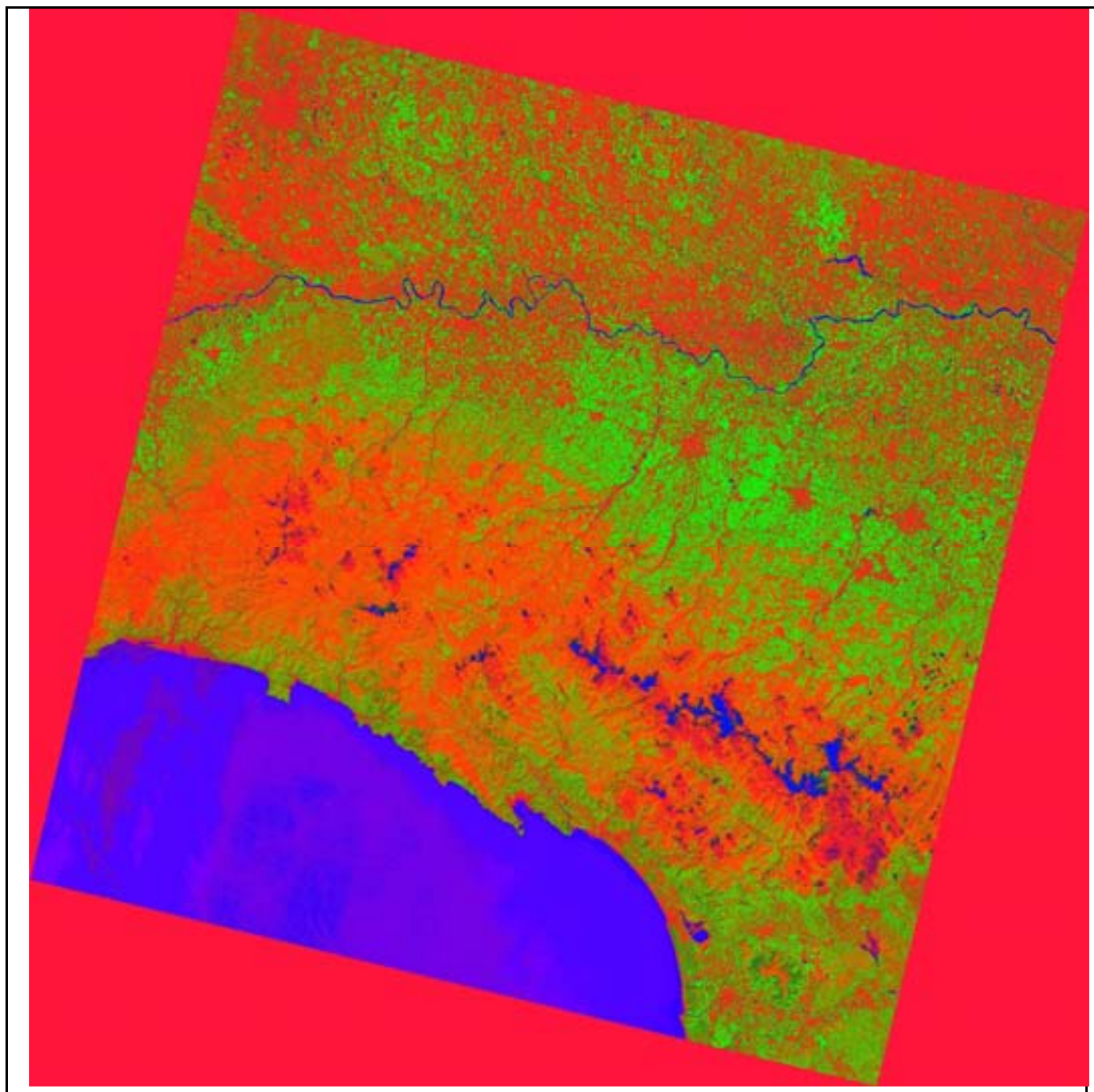

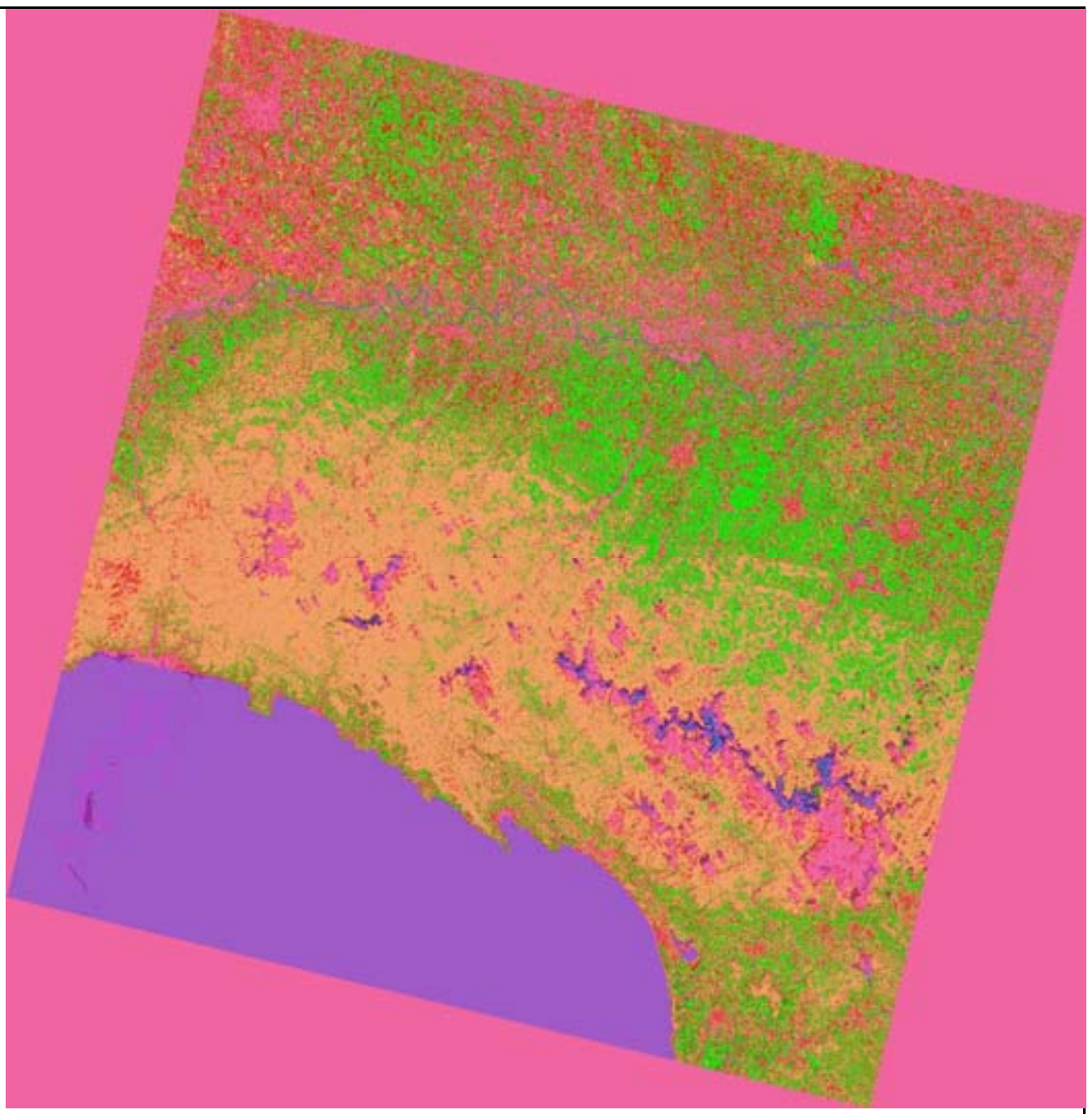

(a) RGB image histograms.

(b) Map histogram and legend.

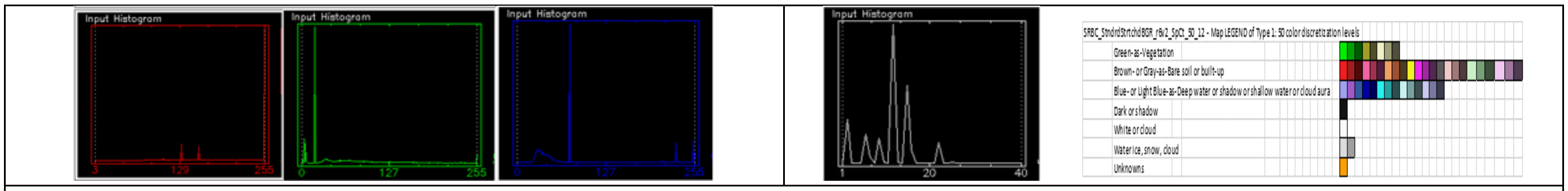

Fig. 12. (a) Test image 5. Landsat-8 OLI+TIRS image, acquired on 2014-04-07, Spatial resolution: 30 m, radiometrically calibrated into TOARF values. Band-ratio spectral indexes (SI), R = Bare Soil SI = Medium Infrared / Near Infrared, G = Vegetation SI = Near Infrared / Visible Red, B = Water SI = Visible Red / Medium Infrared, subject to histogram stretching for color constancy and mounted onto the RGB data space. These three SIs are equivalent to fuzzy membership functions and/or heterogeneous continuous input variables. No stretching is applied for visualization purposes.. (b) RGBIAM-QNQ converter. Color map in pseudocolors, 49+1 color quantization levels.

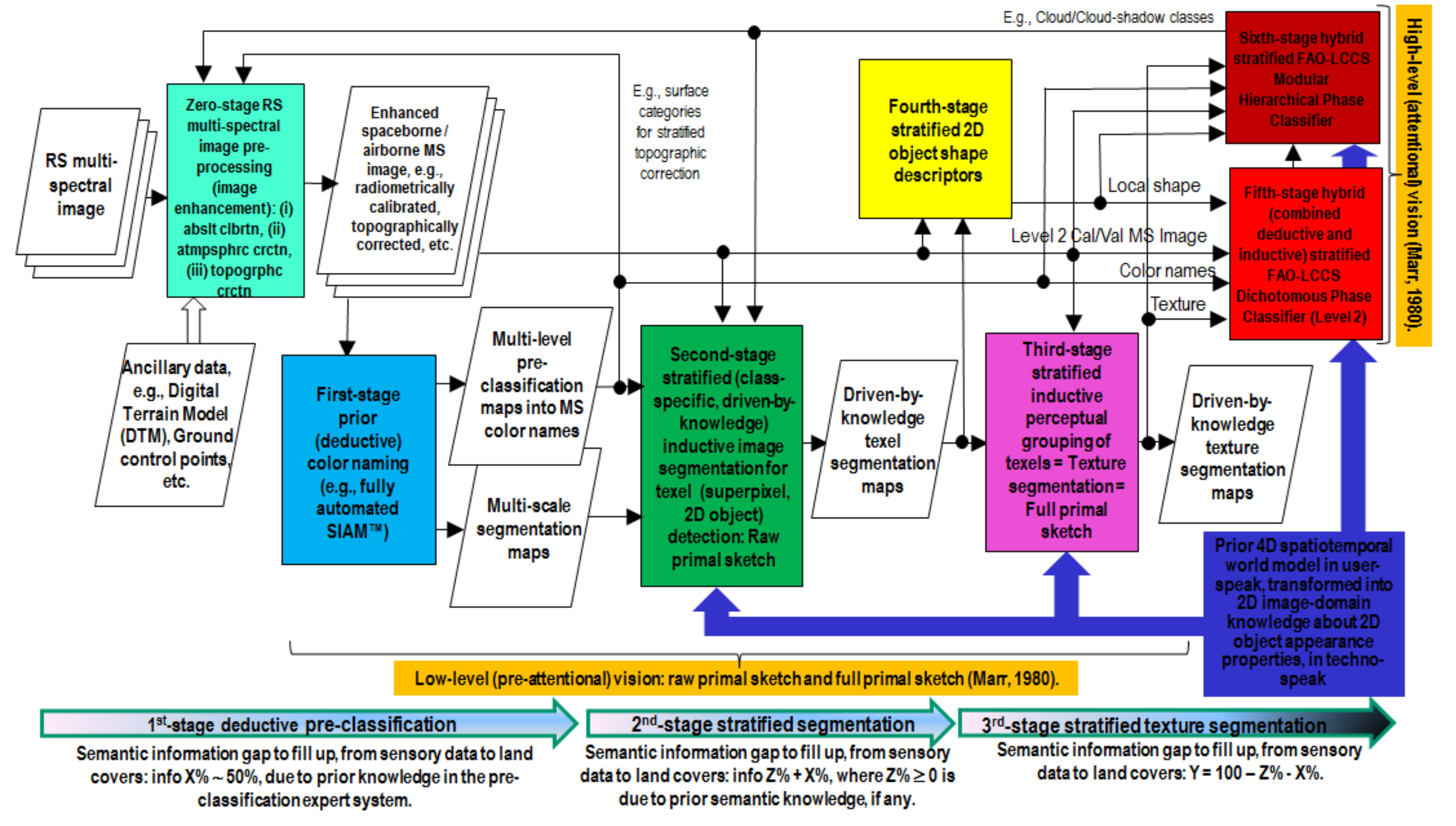

Fig. 13. Novel six-stage hybrid feedback image understanding system (IUS) architecture, proposed to the RS community in recent years. It includes Stage 0 (zero) for stratified image enhancement. According to experiments on EO image classification problems, approximately 50% of the semantic information gap from sensory data to land cover (LC) classes can be filled in by the first-stage expert system for automatic color naming [23], [24].

## TABLES AND TABLE CAPTIONS

TABLE 1. Quantitative comparison of three different realizations of the RGB image QNQ transform, shown in Fig. 4, whose block 2 for VQ is implemented as: (i) the deductive RGBIAM's static decision tree for RGB cube partitioning, (ii) the traditional inductive *k*-means VQ algorithm, where parameters *k* and maximum number of iterations must be user-defined, while the *k* vector centroids are initialized by random input data sampling [35], and (iii) the RGBIAM expert system initializes the inductive *k*-means VQ algorithm. Similar to [54], a three-fold cross-validation is adopted to provide a predictive estimate of the quantization error of the inductive k-means VQ algorithm. One-of-three datasets is employed for training (shown as a gray cell) and the remaining two for testing. In the last row, where pooling of each error indicator or quality index collected across datasets (rows) for each method (column) is accomplished by a sum, best overall results are shown in bold.

| **RMSE, Tot. no. of segments, Mean area, 49 color labels, RGB image** | **Deductive** (one-pass) RGBIAM, 49 + 1 color names | **Inductive** K-means VQ, learning phase in test image 1, k = 49, no. of iterations = | **Inductive** K-means VQ, learning phase in test image 2, k = 49, no. of iterations = | **Inductive** K-means VQ, learning phase in test image 3, k = 49, no. of iterations = | **Hybrid VQ:** first-stage deductive RGBIAM and second-stage | **Mean of the variable across algorithms** | **StDev of the variable across algorithms** |
|---|---|---|---|---|---|---|---|
| Test image 1, Human face, RW = 230, CL = 219. | R: 17.292217, G: 16.468367, B: 15.261817, Tot. no. of segments = 1008, Mean area (pixel) = | R: 12.978549, G: 8.689056, B: 7.933170, Tot. no. of segments = 2123, Mean area (pixel) = | R: 12.183979, G: 8.229305 B: 7.008564, Tot. no of segments = 2949, | R: 11.104136, G: 8.186459 B: 6.905596, Tot. no of segments = 3213, | R: 9.775062, G: 9.010200 B: 9.197877, Tot. no of segments = 1662, Mean area (pixel)= | R: 12.67, G: 10.12, B: 9.26, Tot. no of segments = 2191, Mean area (pixel)= | R: 2.85, G: 3.57, B: 3.48, Tot. no of segments = 908.71, Mean area (pixel)= 13.92 |
| | 49.97 | 23.73 | Mean area (pixel) = 17.08 | Mean area (pixel)= 15.68 | 30.32 | 27.35 | |
| Test image 2, Aerial optical, RW = 2744, CL = 4616. | R: 13.236089, G: 12.725385, B: 12.010231, Tot. no. of segments = 899416, Mean area (pixel) = 14.08 | R: 11.771849, G: 10.646335 B: 10.953172, Tot. no of segments = 1066650, Mean area (pixel) = 11.87 | R: 6.356885, G: 6.009790, B: 5.960453, Tot. no. of segments = 3486527, Mean area (pixel) = 3.27 | R: 10.565770, G: 9.723846 B: 9.675221, Tot. no of segments = 1395591, Mean area (pixel) = 9.07 | R: 5.825976, G: 5.208758 B: 5.193217, Tot. no of segments = 2178923, Mean area (pixel)= 5.81 | R: 9.55, B: 8.86, B: 8.76, Tot. no of segments = 1805421.4, Mean area (pixel)= 8.89 | R: 3.30, G: 3.17, B: 3.032, Tot. no of segments = 1060685, Mean area (pixel)= 4.27 |
| Test image 3, Satellite bi-temporal SAR, RW = 4480, CL = 5012. | R: 4.811451, G: 14.903615, B: 17.178292, Tot. no. of segments = 1859568, Mean area (pixel) = 12.07 | R: 4.957717, G: 23.638669 B: 28.337016, Tot. no of segments = 1192200, Mean area (pixel) = 18.83 | R: 4.839931, G: 23.269672 B: 27.527727, Tot. no of segments = 1379675, Mean area (pixel) = 16.27 | R: 13.713457, G: 10.502449, B: 11.288297, Tot. no. of segments = 9713383, Mean area (pixel) = 2.31 | R: 2.278091, G: 10.321090 B: 12.321693, Tot. no of segments = 3475502, Mean area (pixel)= 6.46 | R: 6.12, G: 16.58, B: 19.33, Tot. no of segments = 3524066, Mean area (pixel)= 11.19 | R: 4.39, G: 6.58, B: 8.17, Tot. no of segments = 3574793, Mean area (pixel)= 6.82 |
| **Total RMSE (to be ↓), Total no. of segments (to be ↓), Total Mean area (to be ↑)** | R = 35.339757, G = 44.097367, B = 44.45034, R+G+B = 123.89, Tot. no. of segments = 2759992, **Tot. Mean area = 76.12** | R = 29.708115, G = 42.97406, B = 47.223358, R+G+B = 119.90, **Tot. no. of segments = 2260973**, Tot. Mean area = 54.43 | R = 23.380795, G = 37.508767, B = 40.496744, R+G+B = 101.38, Tot. no. of segments = 4869151, Tot. Mean area = 36.99 | R = 35.383363, G = 28.434795, B = 27.019478, R+G+B = 91.66, Tot. no. of segments = 11112187, Tot. Mean area = 27.06 | **R = 17.879129, G = 24.540048, B = 26.712787, R+G+B = 69.13**, Tot. no. of segments = 5656087, Tot. Mean area = 42.59 | | |

TABLE 2. Z-scores extracted from Table 1. In the last row, where pooling of each error indicator or quality index collected across datasets (rows) for each method (column) is accomplished by a sum, best overall results are shown in bold.

| **RMSE, Tot. no. of segments, Mean area, 49 color labels, RGB image** | **Deductive** (one-pass) RGBIAM, 49 + 1 color names | **Inductive** K-means VQ, learning phase in test image 1, k = 49, no. of iterations = | **Inductive** K-means VQ, learning phase in test image 2, k = 49, no. of iterations = | **Inductive** K-means VQ, learning phase in test image 3, k = 49, no. of iterations = | **Hybrid VQ:** first-stage deductive RGBIAM and second-stage | **Mean of the standardized variable across algorithms** | **StDev of the standardized variable across algorithms** |
|---|---|---|---|---|---|---|---|
| Test image 1, Human face, RW = 230, CL = 219. | R: 1.62, G: 1.78, B: 1.72, Tot. no. of segments = -1.30, Mean area (pixel) = 1.62 | R: 0.11, G: -0.40, B: -0.38, Tot. no. of segments = -0.07, Mean area (pixel) = -0.26 | R: -0.17, G: -0.53, B: -0.65, Tot. no of segments = 0.83, Mean area (pixel) = -0.74 | R: -0.55, G: -0.54, B: -0.68, Tot. no of segments = 1.12, Mean area (pixel)= -0.84 | R: -1.01, G: -0.31, B: -0.02, Tot. no of segments = -0.58, Mean area (pixel)= 0.21 | R: 0, G: 0, B: 0, Tot. no of segments = -0, Mean area (pixel)= 0 | R: 1, G: 1, B: 1, Tot. no of segments = 1, Mean area (pixel)= 1 |
| Test image 2, Aerial optical, RW = 2744, CL = 4616. | R: 1.11, G: 1.22, B: 1.072, Tot. no. of segments = -0.85, Mean area (pixel) = 1.21 | R: 0.67, G: 0.56, B: 0.72, Tot. no of segments = -0.69, Mean area (pixel) = 0.69 | R: -0.97, G: -0.90, B: -0.92, Tot. no. of segments = 1.58, Mean area (pixel) = -1.23 | R: 0.31, G: 0.27, B: 0.30, Tot. no of segments = -0.39, Mean area (pixel) = 0.042 | R: -1.13, G: -1.15, B: -1.18, Tot. no of segments = 0.35, Mean area (pixel)= -0.72 | R: 0, G: 0, B: 0, Tot. no of segments = -0, Mean area (pixel)= 0 | R: 1, G: 1, B: 1, Tot. no of segments = 1, Mean area (pixel)= 1 |
| Test image 3, Satellite bi-temporal SAR, RW = 4480, CL = 5012. | R: -0.20, G: -0.25, B: 0.26, Tot. no. of segments = -0.46, Mean area (pixel) = 0.13 | R: -0.26, G: 1.08, B: 1.10 Tot. no. of segments = -0.65, Mean area (pixel) = 1.12 | R: -0.29, G: 1.02, B: 1.0, Tot. no of segments = -0.60, Mean area (pixel) = 0.74 | R: 1.73, G: -0.91, B: -0.98, Tot. no. of segments = 1.73, Mean area (pixel) = -1.30 | R: -0.87, G: -0.94, B: -0.86, Tot. no of segments = -0.01, Mean area (pixel)= -0.69 | R: 0, G: 0, B: 0, Tot. no of segments = -0, Mean area (pixel)= 0 | R: 1, G: 1, B: 1, Tot. no of segments = 1, Mean area (pixel)= 1 |
| **Total RMSE (to be ↓), Total no. of segments (to be ↓), Total Mean area (to be ↑)** | R = 2.44, G = 2.75, B = 2.53, **Tot. no. of segments = -2.62, Tot. Mean area = 2.97** | R = 0.52, G = 1.24, B = 1.44, Tot. no. of segments = -1.42, Tot. Mean area = 1.56 | R = -1.43, G = -0.40, B = -0.57, Tot. no. of segments = 1.82, Tot. Mean area = -1.22 | R = 1.49, G = -1.18, B = -1.36, Tot. no. of segments = 2.47, Tot. Mean area = -2.10 | **R = -3.02, G = -2.40, B = -2.05**, Tot. no. of segments = -0.24, Tot. Mean area = -1.20 | | |

TABLE 3. Computation time, including data processing and I/O operations, of the implemented RGB image QNQ transform employing the RGBIAM expert system for VQ. The computational complexity of the RGB image QNQ transform is linear in the image size. Irrespective of I/O operations, computation time complies well with the linear-time hypothesis.

| **Test image, Bands = *B* = 3** | **Rows, *RW*** | **Columns, *CL*** | **Image size in pixels, *N*** | **Computation time in seconds (including I/O operations)** |
|---|---|---|---|---|
| Terrestrial, optical | 219 | 230 | 50370 | 1 |
| Airborne, optical | 2744 | 4616 | 12666304 | 9 |
| Spaceborne, SAR | 4480 | 5012 | 22453760 | 19 |
| | | | | |

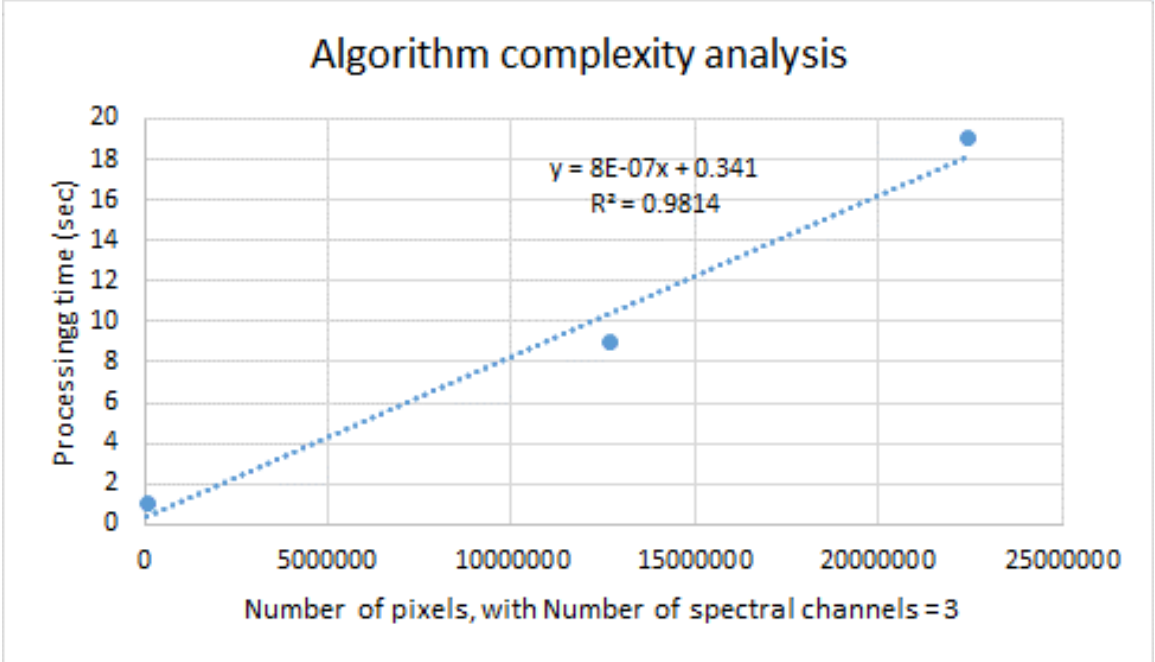